\documentclass[acmsmall]{acmart}

\usepackage{amssymb}
\usepackage{booktabs}
\usepackage{threeparttable}
\usepackage{multirow}
\usepackage{graphicx}
\usepackage{subfigure}
\usepackage{makecell}
\usepackage{url}
\usepackage{caption}
\AtBeginDocument{%
  \providecommand\BibTeX{{%
    \normalfont B\kern-0.5em{\scshape i\kern-0.25em b}\kern-0.8em\TeX}}}

\setcopyright{acmcopyright}
\acmJournal{TOMM}
\acmYear{2021} \acmVolume{1} \acmNumber{1} \acmArticle{1} \acmMonth{1} \acmPrice{15.00}\acmDOI{10.1145/3457893}

\acmJournal{JACM}
\acmVolume{37}
\acmNumber{4}
\acmArticle{111}
\acmMonth{8}

\begin{document}

%%
%% The "title" command has an optional parameter,
%% allowing the author to define a "short title" to be used in page headers.
\title{Health Status Prediction with Local-Global Heterogeneous Behavior Graph}

%%
%% The "author" command and its associated commands are used to define
%% the authors and their affiliations.
%% Of note is the shared affiliation of the first two authors, and the
%% "authornote" and "authornotemark" commands
%% used to denote shared contribution to the research.
\author{Xuan Ma}
\email{maxuan2018@ia.ac.cn}
\author{Xiaoshan Yang}
\email{xiaoshan.yang@nlpr.ia.ac.cn}
\author{Junyu Gao}
\email{gaojunyu2015@ia.ac.cn}
\author{Changsheng Xu}
\email{csxu@nlpr.ia.ac.cn}
\affiliation{%
	\institution{National Laboratory of Pattern Recognition, Institute of Automation, Chinese Academy of Sciences; School of Artificial Intelligence, University of Chinese Academy of Sciences}
	\streetaddress{NO. 95 Zhongguancun East Rd.}
	\state{Beijing}
	\country{China}
	\institution{and Peng Cheng Laboratory}
	\state{Shenzhen}
	\country{China}
}

%%
%% By default, the full list of authors will be used in the page
%% headers. Often, this list is too long, and will overlap
%% other information printed in the page headers. This command allows
%% the author to define a more concise list
%% of authors' names for this purpose.
%\renewcommand{\shortauthors}{Trovato and Tobin%, et al.}

%%
%% The abstract is a short summary of the work to be presented in the
%% article.
\begin{abstract}
   
Health management is getting increasing attention all over the world.
However, existing health management mainly relies on hospital examination and treatment,
which are complicated and untimely.
The emerging of mobile devices provides the possibility to manage people's health
status in a convenient and instant way.
Estimation of health status
can be achieved with various kinds of data streams continuously collected from wearable sensors.
However, these data streams are multi-source and heterogeneous, containing complex temporal structures with local contextual and global temporal aspects, which makes the feature learning and data joint utilization challenging.
We propose to model the behavior-related multi-source data streams with a local-global graph, which contains multiple local context sub-graphs to learn short term local context information with heterogeneous graph neural networks and a global temporal sub-graph to learn long term dependency with self-attention networks.
Then health status is predicted based on the structure-aware representation learned from
the local-global behavior graph.
We take experiments on StudentLife dataset,
and extensive results demonstrate the effectiveness of our proposed model.

%It is noted that students' mental health and acadamic performance are highly relavant to their daily behaviors, including their study habits and lifestyles. A large and comprehensive study StudentLife finds these significant correlations based on the automatic objective sensor data from smartphones and mental health status and educational outcomes of the student body. However, far too little attention to date has been paid to jointly consider the following three observations of the mental health and academic performance: 1)daily health status can also be estimated from the daily sensor data. 2) the final mental health statuses and grades of each student can be comprehensively revealed from the daily statuses. 3) higher semantic concepts can be concluded from the interactions between different medias. In the light of these, we propose an multimedia multi-task learning model to co-regularize the media agreement and temporal progression, to predict the daily health statuses and term mental health and acadamic performance at the same time. Evaluations on the StudentLife dataset well verify our proposed model.

\end{abstract}

%%
%% The code below is generated by the tool at http://dl.acm.org/ccs.cfm.
%% Please copy and paste the code instead of the example below.
%%
%%
%% Keywords. The author(s) should pick words that accurately describe
%% the work being presented. Separate the keywords with commas.

%%
%% This command processes the author and affiliation and title
%% information and builds the first part of the formatted document.

\maketitle

\section{Introduction}
%In 21st century, chronic diseases have become a major threat to people's health. According to the World Health Statistics 2017, about 56 million patients died globally in 2015, of which 40 million deaths were caused by chronic non-communicable diseases, accounting for 70\% of the total deaths \cite{timmurphy}. When compared with other disease, chronic diseases have a long latency period and multiple causes. If we  . For this reason, there is an increasing need for health status estimation.

Health is not only the basic guarantee of human happiness and well-being
but also the foundation of economic progress.
Keeping in good health needs reasonable health management
which has attracted increasing attention from governments and companies over the world.
Existing health management mainly relys on
medical examination and specialized patient treatment in hospitals.
However, many citizens usually do not consider going to the hospital for a check-up until they have abnormal physical symptoms.
Moreover, the regular or irregular medical examination with the professional medical equipment in hospitals can only get some discrete measurements of individual's health status at a specific moment.
%
%Both the two reasons make health status estimation difficult in hospital environment.
%
Both reasons bring a lot of challenges for early detection and prevention of diseases which are the key components of an effective health management.

% add refs
According to a study of World Health Organization(WHO), personal behaviors and lifestyles account for 60\% factors affecting human health \cite{timmurphy}.
For example, non-smokers with defective gene are much less likely to suffer from lung disease than regular smokers \cite{live}.
A healthy diet and adherence to appropriate exercise can greatly reduce the incidence of diabetes and cardiovascular diseases \cite{eating}.
Therefore, the real-time and continuous analysis and monitoring of personal behavior and health status are helpful for individuals to enhance self-active health awareness and learn the disease prevention knowledge, thus improve health management capabilities\cite{min2017you}.
%add refs

%
The rapid development of smart portable and wearable devices has promoted the widespread use of various low-power sensors, and the advent of the 5G era has made it possible to collect individual health data streams of multiple sensors in real time.
%
%Using mobile devices such as smartphone and smartwatch to do the health status estimation and health management is becoming more and more popular.
%
As long as people carry their devices, all their daily routines, diets and activities are recorded automatically and instantly without extra efforts.
For example, the daily routine can be recorded by the GPS sensor, a picture showing people's diet can be captured using phone camera, while the activity information can be recognized using the accelerometer sensor.
These real time data streams can be transferred to a back-end system for
behavior analysis and health status estimation
and finally help people improve their health.
Compared with the patient records in hospital,
this kind of data streams not only provide long-term signals to fully describe the individual's daily behavior and lifestyle,
but also support continuous data transmission and analysis without interfering with individual's daily lives and work.

The health data streams collected from various sensors are multi-source and heterogeneous.
On the one hand,
the sampling frequency of each sensor is different, which makes the co-processing difficult.
On the other hand, the collected sensor data are multimodal,
\textit{e.g.} pictures or videos from cameras, motion signals from accelerometers and  gyroscopes, and location coordinates from GPS sensors.
Though, multi-source sensors can provide complementary information,
the feature learning and joint utilization of the multimodal data streams
remain a challenge for health status estimation.
%{\color{red}behavior analysis and health status estimation.}
%the various sources of these data n\right) ot only bring differences in sampling frequency, but also cause the modal di, which pose challenges on data co-precessing and joint utilization.

%
There have been a bunch of health status prediction methods based on mobile devices.
%{\color{red}
A few methods rely on non-parametric methods (\textit{e.g.}K-Means, Mean Shift)
with the single-source data as input,
such as acceleration signal \cite{natale2012monitoring}, camera data in smartphone\cite{banhalmi2018analysis}, sound from the microphone\cite{nandakumar2015contactless}.
Other methods are devoted to taking advantage of information from various sources\cite{min2014toss}\cite{li2019deep}.
%}
%%%%ÉÏÃæ±êºìµÄÕâ¶ÎÊÇ¶ÔÒÑÓÐ·½·¨×Ü½á£¬¼ì²éÒ»ÏÂÊÇ²»ÊÇ×Ü½áµÄºÏÀí£¬¿ÉÒÔµ÷ÕûºÍ²¹³ä£»
However,
most of exiting methods do not make full use of the
structure information of the multi-source data streams,
which affect the further improvement of performance.
%
%Thus,
%to perform health status estimation,
%there is a need to utilize the multi-source and heterogeneous health data stream of individual behaviors in a reasonable way.
%
Practically, the multi-source and heterogeneous health data stream
mainly reflects the individual behaviors and lifestyles
which contain complex temporal structures with local contextual and global temporal aspects.
Local context refers to the behavior in a short term, such as the activities and routines in one day.
Detailed behavior information such as the activity sequence and location transfer should be considered in local context to get the principal characteristic of the daily behavior.
While for the global temporal aspect, the temporal dependency
among local contexts need to be captured for representing
the long-term comprehensive description of individual behaviors.
%

%From this perspective, the multi-source and heterogeneous health data are structural both for its local and global characteristics, in which case the structural networks are usually used.
%

%
Recently, Graph Neural Networks (GNNs)\cite{wu2020comprehensive}\cite{zhang2020deep}\cite{gao2018watch} have drawn great attention in modeling interactions in structural data. Taking a graph as input, GNNs propagate messages between nodes according to the edges, thus learn the representations for both nodes and edges.
Most GNNs work on the homogeneous cases where the nodes in graph belong to one type \cite{atwood2016diffusion}\cite{niepert2016learning}\cite{gilmer2017neural}.
Heterogeneous graph neural network\cite{sun2013mining},
a special case for GNN,
is devoted to solving the other situation where nodes are of different types.
It has been successfully applied in \cite{dong2017metapath2vec}\cite{chen2018pme}\cite{wang2019heterogeneous}
where highly competitive performances are obtained.
Inspired by its development,
it is promising to model the intra-modality structure
and inter-modality interaction of the multi-source and heterogeneous health data
with the heterogeneous graph neural networks.
In this paper, we propose to predict the daily mental health status based on the multi-source wearable sensor data. 
We build a local-global individual behavior graph (LGIBG) based on the heterogeneous data,
and then predict the daily health status with the help of heterogeneous graph neural networks. 
Specifically, we take 3 kinds of sensor data streams (accelerometer, audio, wifi) as input, and detect middle-level behavior-related concepts (\textit{i.e. }walking, running, silence)
with pretained backbone models.
These concepts are further used to build the local-global individual behavior graph
which consists of multiple local context sub-graphs and a global temporal sub-graph.
The local context sub-graphs are created with the concepts detected from daily data streams as heterogeneous nodes
%with these concepts as heterogeneous nodes
which are connected with homogeneous and heterogeneous edges.
Next, a densely connected global temporal sub-graph is created on top of the local context sub-graphs.
Then we take advantage of heterogeneous neural network to learn the features of local context sub-graphs and get both the semantic and structural representations.
The representation of the global temporal sub-graph is learned with a self-attention network,
and it is finally used to predict the health status.

In summary, the contributions of our work are threefold:
%\begin{itemize}
% {\color{red}
\textbf{(1)} To effectively represent the behavior-related multi-source data collected from
wearable sensors, we build a local-global graph which consists of multiple local context sub-graphs and a global temporal sub-graph.
The local-global graph can well describe the short-term context information of individual behaviors and their long term temporal dependencies.
\textbf{(2)} We learn the short term semantic and structural representations from local context sub-graphs with heterogeneous graph neural networks,
%on local context graphs to get the short term information and .
and the long term representation from the global temporal sub-graph with self-attention networks.
\textbf{(3)} We demonstrate the effectiveness of the proposed method in health prediction on public dataset Studentlife.

\section{Related Work}

\subsection{Health Status Prediction}
% {\color{red}
Our task is to predict the health status based on personal behaviors with the multi-source sensor data collected in daily life. 
This task has important medical implications because it can provide early prevention of diseases and complement the clinical treatment in hospital.
%
%which 
%
As we know, most existing health prediction methods can be divided into two categories: Electronic Health Record(EHR) based methods~\cite{{mcgrath2013toward,mwangi2012prediction}} and mobile sensor data based methods~\cite{canzian2015trajectories,burns2011harnessing}. 
The EHR based methods are more relevant to medical studies since they use the record data collected during hospital treatment process with professional medical equipment.
However, people have few EHR data before they are detected with diseases, thus this kind of methods, however good, can only provide a piece of the puzzle.
%%%YXS: check the description
%
More kinds of data about early-life experience of patients should also be considered in treatment, which is actually what the second kind of methods do.
The sensor data based methods focus on the personal behavior in daily life and take advantage of mobile devices, which are more convenient to monitor people's health and also provide additional useful information for clinical treatment. Below we will introduce these two categories in detail.

% The health status prediction task aims to provide a quantitative evaluation of person's health status, which can be used to predict and prevent the diseases. There are two major directions towards the health status prediction: Electronic Health Record(EHR) based methods and mobile sensor data based methods.

Electronic Health Record (EHR) data are collected during the hospital treatment process, containing nearly all the information of patients, such as the diagnoses, medication prescriptions and clinical notes.
In early stages, expert-defined rules are adopted to identify disease based on the EHR data, such as type 2 diabetes\cite{kho2012use} and cataract \cite{peissig2012importance}.
%{\color{red}Add several methods using shallow methods...XXX.}
Much work based on EHR data has been done with deep learning models, with disease classification task most commonly. Cheng \textit{et al.} \cite{cheng2016risk} and Acharya \textit{et al.} \cite{acharya2018deep} train a CNN model to classify the normal, preictal, and seizure EEG signals. Che \textit{et al.} \cite{che2017rnn} propose a multi-class classification task to predict the different stages of Parkinson's disease with Recurrent Neural network(RNN).
Kam \textit{et al.} \cite{kam2017learning} do binary classification of sepsis by regarding the EHR data as input to a long short-term memory network (LSTM).
In addition to the disease classification,
the future event prediction is another task attracting much attention recently,
which aims to predict the future medical events according to the historical records. For example, Joseph \textit{et al.} \cite{futoma2015comparison} and Rajkomar \textit{et al.} \cite{rajkomar1801scalable} use the EHR data from the hospital
to predict events such as mortality, readmission, length of stay and discharge diagnoses
with deep feed forward network and LSTMs.
% {\color{red}
There has been much work on mental health prediction based on EHR data, among which T1-weighted imaging~\cite{mcgrath2013toward,mwangi2012prediction} and functional MRI (fMRI)~\cite{fu2008pattern,hahn2011integrating} are the most commonly used data to study brain structure, with other physiological signals such as electroencephalogram also playing an important role.
Recently,
machine learning receives
more attention for its effect on improving the management of mental health.
Costafreda \textit{et al.} \cite{costafreda2009prognostic} do a depression classification task with SVM using the smoothed gray matter voxel-based intensity values. 
Rosa \textit{et al.} \cite{rosa2015sparse} propose a sparse L1-norm SVM to predict depression with the feature of region-based functional connectivity. 
Cai \textit{et al.} \cite{cai2018pervasive} collect the electroencephalogram (EEG) signals of participants and use four classification methods (SVM, KNN, DT, and ANN) to distinguish the depressed participants from normal controls.
% }

%
%However, predicting disease or health status from EHR data has a high demand for the detailed information of symptoms. For example, the above depression prediction related works mainly rely on structural magnetic resonance imaging(sMRI). This brings difficulties in real applications, since people have little EHR data before they are detected with diseases. So these methods, as said in \cite{kim2018application}, however good, con only provide a piece of puzzle. More kinds of data, especially early-life experience of patients also should be considered in treatment. 
% However, predicting disease or health status from EHR data has a high demand for the detailed information of symptoms. For example, the depression prediction related work on EHR data mainly relies on structural magnetic resonance imaging(sMRI). This brings difficulties in real applications, since people have little EHR data before they are detected with diseases. So these methods, as said in \cite{kim2018application}, however good, con only provide a piece of puzzle. More kinds of data, especially early-life experience of patients also should be considered in treatment. To this end, some work has been done on the sensor data based health prediction. 

% Compared with the study which mostly rely on medical data and work after diagnosed, the sensor data based methods focus on the personal behavior in daily life and take advantage of mobile devices, which is more convenient to monitor people's health and also provide assistance to clinical treatment. }
Mobile devices provide another way for health status prediction, where diverse sensors can be used to catch various signals of people, thus make it easier to monitor daily behavior and predict health status.
Machado \textit{et al.} \cite{machado2015human} calculate the signal magnitude area of the acceleration signal to recognize activities with several cluster algorithms(eg. K-Means, Mean Shift).
Koenig \textit{et al.} \cite{koenig2016validation} and Banhalmi \textit{et al.} \cite{banhalmi2018analysis} use camera in smartphone to monitor the heart rate(HR) and heart rate variability(HRV), which are vital signs of cardiovascular health.
Stafford \textit{et al.} \cite{stafford2016flappy} and Goel \textit{et al.} \cite{goel2016spirocall} detect the sound of breathing and cough by microphone in smartphone for assessing pulmonary health in a quick and efficient way.
Besides the use of data from only one source,
many researchers are devoted to taking advantage of information from various sources\cite{huang2020knowledge}\cite{zhang2019multimodal}\cite{qi2020emotion}. Asselbergs \textit{et al.} \cite{asselbergs2016mobile} integrate accelerometer data, call history as well as the short message service pattern to predict the mood. Burns \textit{et al.} \cite{burns2011harnessing} predict depression based on GPS, accelerometer and light sensor data from smartphone. 
Nag \textit{et al.} \cite{nag2018cross} estimate the heart health status by combining sensor data from wearable devices and other factors, such as inherent genetic traits, circadian rhythm and living environmental risks analysed from cross-modal data, which provides better personalized health insight.

However, all of the above work cannot well explore local and global temporal characteristics of the daily behavior based on multi-source wearable sensors.

\subsection{Graph Neural Network}

Recently, the emergence of structural data, especially the structured graphs promote the development of Graph Neural Networks(GNNs) \cite{gao2019graph}\cite{wu2020comprehensive}\cite{zhang2020deep}.
As the early work of GNNs, recurrent graph neural networks (RecGNNs) \cite{scarselli2008graph}\cite{gallicchio2010graph} apply recurrent architectures to learn the node representation, where message passing is done constantly with nodes' neighborhoods until the node representations are stable.
Inspired by the success of Convolutional Neural Network(CNN),
\textit{convolution} operation is also introduced to graph data in both spectural\cite{henaff2015deep}\cite{defferrard2016convolutional}\cite{kipf2016semi} and spatial ways\cite{atwood2016diffusion}\cite{junyu2019AAAI_TS-GCN}\cite{gao2020learning}.
The spectral approaches adapt the spectral graph theory to design a graph convolution.
The spatial approaches inherit the message passing idea in RecGNNs while have the difference in getting node representations by stacking multiple convolutional layers.
Besides RecGNNs and ConvGNNs, many other graph architectures have been developed to cope with different scenarios.
For example, Graph autoencoders(GAEs)~\cite{cao2016deep}\cite{wang2016structural} are used to learn the graph embedding by reconstructing the structural information such as adjacency matrix of graph.
Spatial-temporal graph neural networks (STGNNs) \cite{seo2018structured}\cite{li2017diffusion}\cite{jain2016structural} aim to model both the spatial and temporal dependency of data and learn the representation of spatial-temporal graph, which have advantages in the related tasks, such as human action recognition.

%·Ö¿ªÐ´ÄÇÐ©¹¤×÷µÄÈ±µã£¿£¿ »òÕßÍ»³öËûÃÇÃ»ÓÐÀûÓÃ²»Í¬ÀàÐÍ½ÚµãÖ®¼äµÄ¹ØÏµ¡£
Most of the existing GNNs focus on homogeneous graphs where nodes are in the same type and can be calculated in the same way.
In comparison, heterogeneous graph contains diverse types of nodes and edges, leading to a more complicated situation in calculation.
On the one hand, different types of nodes may have different semantic meanings and in different feature spaces.
On the other hand, the heterogeneous graph represents both homogeneous and heterogeneous relations of data.
Recently, some work has been done on heterogeneous graphs.
Dong \textit{et al.} \cite{dong2017metapath2vec} propose a path2vec method to learn heterogeneous graph embeddings with a meta-path based random walk. Chen \textit{et al.} \cite{chen2018pme} process different kinds of nodes with several projection matrices used to embed all the nodes into a same space and then do the link prediction.
Wang \textit{et al.} \cite{wang2019heterogeneous} further introduce the hierarchical attention to heterogeneous graph to learn attentions for both nodes and meta-paths.
Until now, the application
of heterogeneous GNNs to individual behavior analysis and health status prediction is yet to be explored.
%
%{\color{red}
%Moreover, existing methods
%either transform all kinds of nodes into a common space
%or capture heterogeneous relations by extracting multiple adjacency matrices according to the heterogeneous neighborhoods.
%These methods actually work on homogeneous nodes
%and do not consider the direct interaction between heterogenous nodes.
%%
%In comparison, our method processes the heterogenous nodes in their separate spaces and models their interactions directly instead of using the adjacency matrixes,
%thus obtains more semantic information and structural information of different kinds of nodes.
%%
%}
%%%%ÉÏÃæÕâ¶Î±êºìµÄ»°Òª¼ì²é£¬Õâ¸öÓëÒÑÓÐµÄGNN·½·¨Çø±ð£¬ÒªÓÐÊµÑéÖ§³Ö£¬·ñÔòÈÝÒ×±»argue

\begin{figure*}[htbp]
	\centering
	%\vspace{-0.1cm}
	%\setlength{\abovecaptionskip}{-0.02cm}
	\setlength{\belowcaptionskip}{-0.2cm}
			\includegraphics[width=14cm, height=4.5cm]{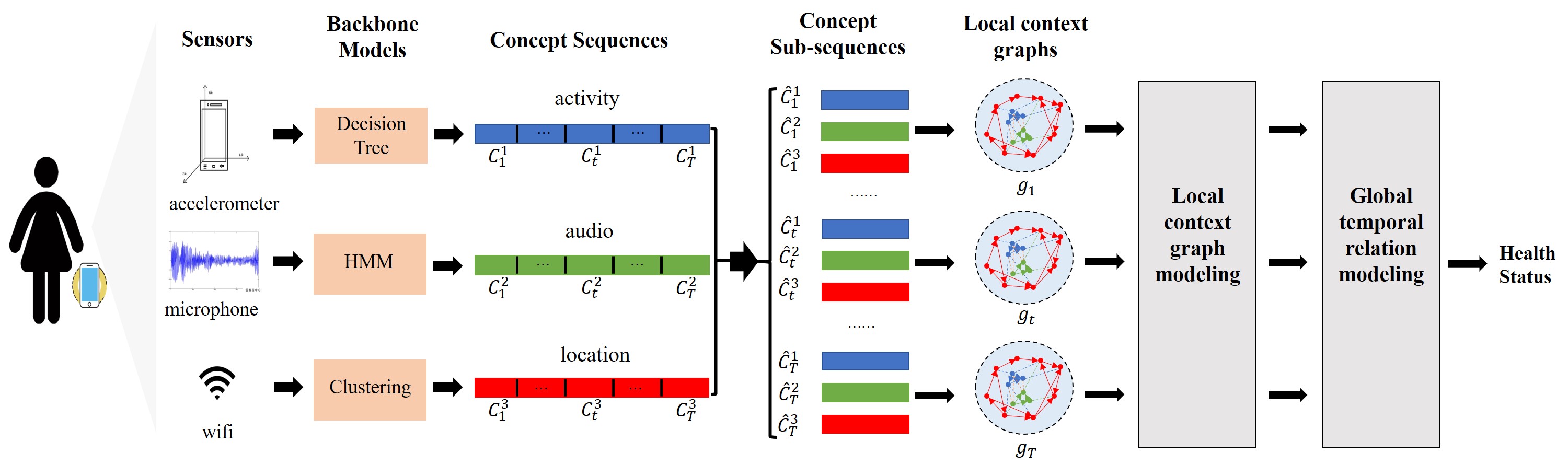}
	\centering
	\caption{Overview of our framework. We take multi-source data streams as input, and detect concept sequences to build a behavior graph which consists of local context sub-graph and global temporal sub-graph.}
	\label{fig:framework1}
\end{figure*}

%The semantic and structural representation of the local context graph are learned with heterogeneous graph neural network. Then structure-aware representation of the global temporal graph is also learned to predict the health status.

\section{Methods}\label{sec:method}

\subsection{Framework Overview}
The individual behavior refers to the way that a person lives.
Our purpose is to model the individual behavior in a period based on multi-modal data streams collected by wearable devices,
and then learn effective representations to predict the health status.

As shown in Figure \ref{fig:framework1}, we take multi-source data streams as input,
and detect behavior-related middle-level concept sequences with pre-trained backbone models.
Then, the behavior-related middle-level concept sequences are used to build the behavior graph
which consists of multiple local context sub-graphs and a global temporal sub-graph.
Specifically, the concepts are regarded as different types of nodes to build local context sub-graphs.
Each local context graph is regarded as a node in the global temporal sub-graph to catch temporal dependency.
%
%As shown in Figure \ref{fig:framework},
%we take multi-source data streams as input,
%and detect behavior-related middle-level concept sequences with pre-trained %backbone models.
%
%{\color{red}
The representations of local context sub-graph and global temporal sub-graph are learned by local context modeling and global temporal relation modeling, based on which the final representation of the behavior graph is learned and used to predict the health status.
%}
%The detailed process of local context graph modeling and global temporal relation modeling is shown in \ref{fig:framework2}.For the local context graph learning,we use the node passage passing module to learn node embeddings and the edge embedding learning module to get the edge embeddings.
%
%Then the graph semantic representation and structural representation are computed with the learned node embeddings and edge embeddings.
%
%For the global temporal graph learning,
%we take the local context graph representations as input, and take advantage of self-attention network to get the final structure-aware representation of the global temporal graph, which is then used to predict the health status.

\subsection{Behavior-related Concept Detection}\label{subsec:backbone}
We take three kinds of data streams (i.e. accelerometer, microphone and wifi) as input,
and detect behavior-related middle-level concept sequences $\{C^1, C^2, C^3\}$ with three pre-trained backbone models (i.e. \textit{activity, audio, location} detectors),
where each concept sequence $C^k = \{c^k_1, c^k_2, \cdots c^k_T\}$
and the $c_k^t$ denotes the specific concept class (e.g. $walking$ detected by the $activity$ detector).
Meanwhile, we also obtain timestamp sequences $\{U^1,U^2,U^3\}$,
where $U^k = \{u^k_1, u^k_2, \cdots u^k_T\}$
%
%that represent the sampling time,
and each timestamp $u^k_t$ is a 2-dimensional vector which represents the start and end time of the detected concept class $c^k_t$ in the corresponding data stream.
More details of the pre-trained backbone models are introduced in Section~\ref{subsec:impl}, and the notations and their corresponding explanations are shown in Table ~\ref{tab:notation}.
%
%{\color{red}Add details of backbone models.XXX}
%
\begin{table}\footnotesize
	\caption{Notations and explanations}
	%\setlength{\abovecaptionskip}{-0.1cm}
	%\setlength{\belowcaptionskip}{-0.9cm}
	%\vspace{-0.3cm}
	\begin{tabular}{cc}
		
		\toprule
		Notation&Explanation\\
		\midrule
		$c^k_t$& concept class for type $k$ at the $t$ th moment\\
		$u^k_t$& a 2-d vector of start and end time of $c^k_i$\\
		%{\color{red}$k_i$} & {\color{red}the $i$th node in the $k$th type.} \\
		$a^k_i$& the attribute of $i$-th node from type $k$.\\
		$x^k_i$& the embedding of $i$-th node from type $k$.\\
		$w_{ij}$& the weight of edge between node pair $(i,j)$,\\
		%		$w'_{ij}$& \makecell[c]{the weight of homogeneous edge between node pair $(i,j)$,\\ while $i, j$ belong to different types }\\
		$e_{ij}$& the embedding of edge between node pair $(i,j)$\\
		%		$e_{k_ik'_j}$& the embedding of heterogeneous edge between node $\mathbb{N}_{k_i}$ and $N_{k'_j}$\\
		\bottomrule
		
	\end{tabular}
	\label{tab:notation}
	%\vspace{-0.1cm}	
\end{table}
\subsection{Behavior Graph Building}
To capture the temporal structure of the individual behavior,
we need to build a behavior graph that contains both the local context information and the long-term relationships from the multimodal data stream.
However, a huge densely connected network may increase the computation complexity and impact the performance.
For this reason, we decompose the whole graph into two kinds of sub-graphs: local context graphs to explore the local information of individual behaviors in a short term, and a global temporal graph to capture temporal dependency in the long term.
The local context graphs are regarded as nodes of the global temporal graph.
%as shown in Figure~\ref{fig:framework}.

\subsubsection{Local Context Sub-Graph}\label{subsubsec:localgraph}
The local context graph is built based on the daily data streams, which could reflect the individual behaviors (e.g. activities, audio, locations) from various aspects.
Taking these individual behaviors into consideration, the local context graph is actually a heterogeneous graph.
%
%Then we split the concept sequences as well as the timestamp sequences into $T$ segments.
%
As illustrated in Section~\ref{subsec:backbone},
we have detected three types of concept sequences $\{C^1, C^2, C^3\}$
from the data streams.
%
%{\color{red}
In the following, we explicitly use $activity$, $audio$, and $location$ to denote the type names.
%}
%
For the $t$-th time step, we use a sliding time window of one day
to crop out three concept sub-sequences $\{\hat{C}^1, \hat{C}^2, \hat{C}^3\}$ from the original concept sequences.
Each concept sub-sequence $\hat{C}^k=\{c^k_{t-h_k/2}, \cdots, c^k_t, \cdots, c^k_{t+h_k/2}\}$ represents
several consecutive concepts detected in one day.
Here the $h_k$ denotes the size of the sliding time window which represents the number of timestamps
contained in one day for the $k$-th type of the concept sequence.
Accordingly, we can obtain the timestamp sub-sequences $\{\hat{U}^1, \hat{U}^2, \hat{U}^3\}$,
where $\hat{U}_k=\{u^k_{t-h_k/2}, \cdots, u^k_t, \cdots, u^k_{t+h_k/2}\}$.
Next, we will introduce how to create the local context graph based on $\{\hat{C}^1, \hat{C}^2, \hat{C}^3\}$ and $\{\hat{U}^1, \hat{U}^2, \hat{U}^3\}$.
For the $t$-th time step,
the local context graph can be formally defined as $\mathcal{G}_t=(\mathcal{V}_t, \mathcal{E}_t)$, where $\mathcal{V}_t = ({\mathcal{N}_1, \mathcal{N}_2,\mathcal{N}_3})$, representing different types of nodes.
Each type corresponds to a specific aspect to describe individual behavior.
%and $K$ is the number of types.
%
$\mathcal{E}_t$ is the set of edges, containing both the homogeneous edges, which connect two nodes in the same type and heterogeneous edges which connect two nodes in different types.

%For the detailed local context graph building process, we choose three representative sensors: accelerometer, microphone and bluetooth to record the individual states. We take their data streams in one day as the input to backbone modules and get three time sequences with different types of middle-level semantic concepts of (\textit{activity, audio, location}). Then we regard these concepts as three types of nodes and build the graph based on these time sequences. Note that each node type has many subtypes indicating a specific state of individual, such as \textit{running} in \textit{activity}, \textit{silence} in \textit{audio}, \textit{library} in \textit{location}.

%{\color{red}
The nodes of the local context graph are comprised of all the concept classes in $\{\hat{C}^1, \hat{C}^2, \hat{C}^3\}$.
%}
%
Since there are three types of concept classes,
the local context graph has three different node types
and thus is a heterogeneous graph.
Each node has an attribute and an embedding representation,
noted as $a_i^k$ and $x_i^k$ for the $i$-th node of type $k$.
For the attribute $a_i^k$ of the node with concept class $c^k_i$,
we use its corresponding timestamp $u^k_i$ to compute the time interval which represents
duration of the concept-related behavior.
For the node embedding $x_i^k$, we introduce external semantic knowledge to help its learning.
Specifically,
we extract Glove embeddings\cite{pennington2014glove} corresponding to the concept class name of each node,
which are pre-trained on Wikipedia according to the word by word co-occurrence.
The embeddings are proved effective in capturing semantic meanings of words in many NLP tasks.
Therefore, these word embeddings would provide a reasonable representation for nodes at first.

As for edges, homogeneous edges and heterogeneous edges are considered in different ways.
%
%{\color{red}
Two nodes with the same type $k$ are connected with homogeneous edges if they are temporal neighborhoods in the concept sequence $\hat{C}^k$.
For example,
a node with concept class ${c}^3_t = dormitory$ from the type $loation$ is connected to
another node with concept class ${c}^3_{t+1} = library$ from the same type if
an individual moves to library from dormitory with continuous timestamps in reality.
%}%%%¼ì²éÃèÊöÊÇ·ñ×¼È·
%
The weight of the homogeneous edge is noted as the frequency of two nodes being neighborhood in the concept sequence.
With the edge weight,
the homogeneous edge captures specific patterns of an individual's behavior change information.
%
%{\color{red}
While for heterogeneous nodes (e.g. $c^k_i$, $c^{k'}_j$, $k\neq k'$),
we connect them according to their co-occurrences in time, i.e., $u^k_i \cap u^{k'}_j \neq \varnothing$.
For example, a node with concept class ${c}^2_t = silence$ from the type \textit{audio} and a node with concept class ${c}^3_t = library$ from the type \textit{location} are connected when they
are detected from the data streams at the same timestamp.
%}%%%%¼ì²éÃèÊöÊÇ·ñ×¼È·
%are recorded with the same timestamp in their separate concept sequences.
%
The co-occurrences in time reflect the interactions of heterogenous nodes,
which could describe the individual behaviors from more aspects.
We do not connect heterogeneous nodes which are temporal neighbouring in the concept sequence, because connecting different types of nodes according to their temporal relations has no practical significance.
The weight of the heterogeneous edge is the frequency of co-occurrences in time.
Note that the weights of both homogeneous edges and heterogeneous edges are written as $w_{ij}$ with $i,j$ as node indexes.
Whether $w_{ij}$ represents a homogeneous edge or a heterogeneous edge depends on specific types of node $i$ and node $j$.

\subsubsection{Global Temporal Sub-Graph}\label{subsubsec:globalgraph}
The global temporal sub-graph models the long-term time dependency of the daily information and gets the global information for the whole period, which is used to predict the health status.
Formally,
the global temporal graph is denoted as $\mathcal{G}=(\mathcal{V}, \mathcal{E})$,
where nodes in $\mathcal{V}$ refer to local context graphs introduced in Section~\ref{subsubsec:localgraph}, and $\mathcal{E}$ represents interactions between any two local context graphs.
%

%\subsection{Graph neural network on spatial-temporal individual lifestyle graph}

\subsection{Local Context Graph Modeling}\label{subsec:localgraphmodel}
\begin{figure*}[t]
	\centering
	%\vspace{-0.1cm}
	%\setlength{\abovecaptionskip}{-0.02cm}
	\setlength{\belowcaptionskip}{-0.2cm}
	\includegraphics[width=11cm]{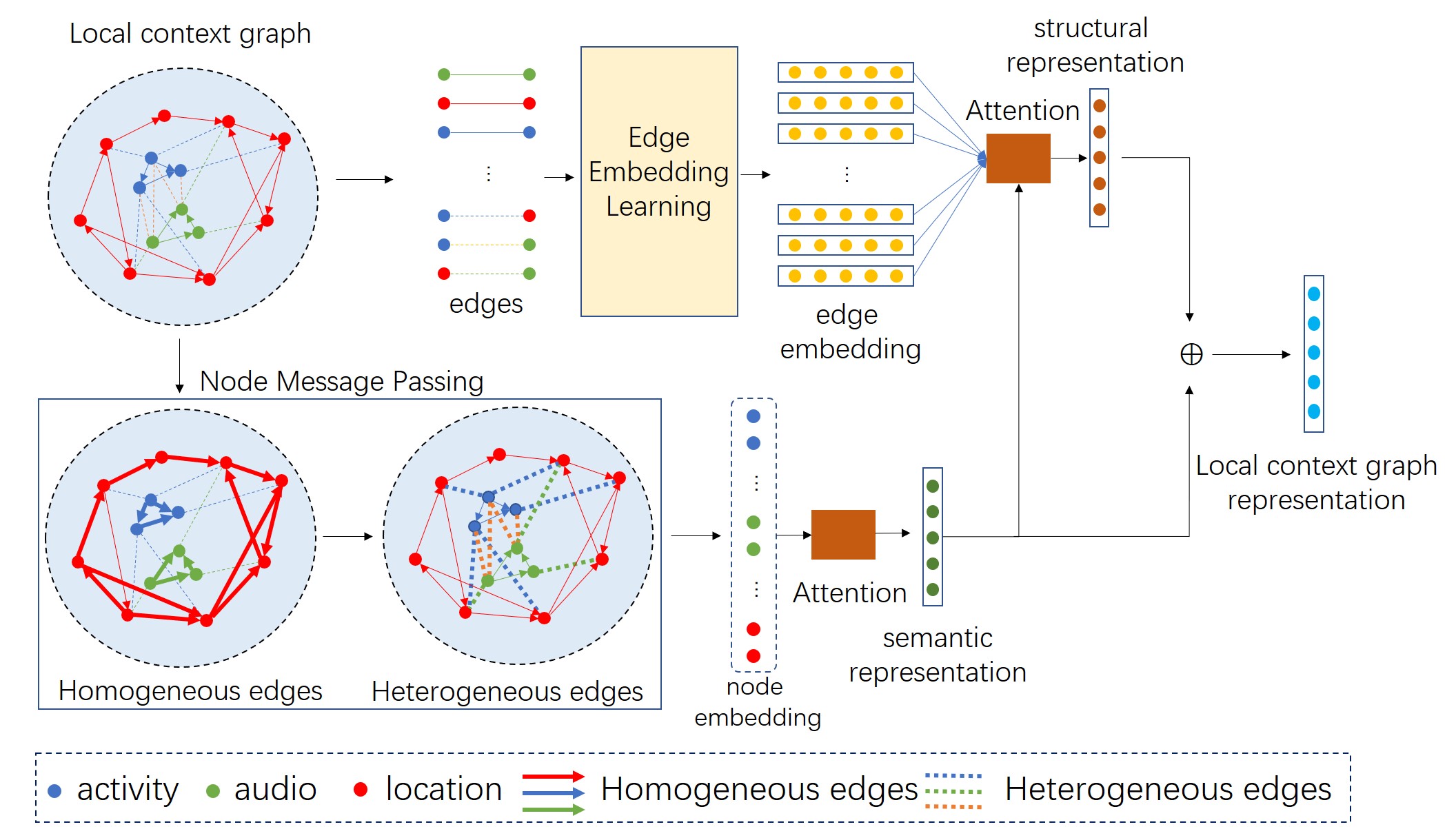}
	\centering
	\caption{Local context graph modeling.}
	\label{fig:framework2}
\end{figure*}

%Now we will introduce
%.
%
As introduced in Section~\ref{subsubsec:localgraph},
we have created a heterogeneous local context graph $\mathcal{G}_t$ at a each time step of multi-source data streams.
%
%Since our concept sequences are divided into $T$ segments and local context graph is built for each segment
%\color{red}{
Now we will introduce how to capture local context information of the short term individual behavior with the heterogeneous graph neural network.
The network contains $m$ layers of node message passing modules and edge embedding learning modules.
%
%{\color{red}
Here we only introduce these two kinds of modules for one layer.
%As shown in \ref{fig:framework2}, we use the node passage passing module to learn node embeddings and the edge embedding learning module to get the edge embeddings.
%
%Then the graph semantic representation and structural representation are computed with the learned node embeddings and edge embeddings, and the final representation of local context graph is got with the combination of semantic and structural representation.
As shown in Figure \ref{fig:framework2}, the node message passing module is used to learn the node embeddings and graph semantic representation, while the edge embedding learning module is used to learn the edge embeddings and  graph structural representation.
Then the final representation of the local context graph is obtained with the combination of
the semantic and the structural representation.
%}
It is worth noting that all local context graphs share the same network parameters.

\subsubsection{Node Message Passing}\label{subsubsec:nodemessage}
We consider the node message passing process in two ways:
homogeneous message passing through homogeneous edges and heterogeneous message passing through heterogeneous edges.
At first, we multiply each node embedding $x_i^k$ with its attribute $a_i^k$, which reveal the node importance in time.
For simplicity, the representation of each node is still denoted by $x_i^k$.

Homogeneous message passing aims to learn information from the same type of nodes according to their edges.
For the \textit{i}th node of type $k$, its message passing process is done as below:

\begin{center}
%\vspace{-0.2cm}
	\begin{equation}
	x^k_i = W^k_x x^k_i + W^k_{\alpha}\sum_{j, j\neq i}\alpha_{ij}x^k_j
	\end{equation}
%\vspace{-0.2cm}
\end{center}
where $W^k_x$ and $W^k_{\alpha}$ are learnable matrices.
The $i$ and $j$ are node indexes, and $\alpha_{ij} = \frac{w_{ij}}{\sum_{j,j\neq i}{w_{ij}}}$ is the normalized value of the homogeneous edge weight between the $i$-th node and the $j$-th node defined in Section~\ref{subsubsec:localgraph}.
The calculations for other types of nodes are in the same way with different projection matrices.
By this means, each node gets information from its homogeneous neighborhoods according to their connections.

Heterogeneous message passing manages to capture additional semantic meanings from other types of nodes,
thus learns a comprehensive representation for individual behaviors.
Since each node has more than one type of heterogeneous neighbours,
we add embeddings learned from all types of heterogeneous neighbors to do the message passing,
shown as follows:
\begin{center}
%\vspace{-0.2cm}
	\begin{equation}
	\setlength{\abovedisplayskip}{-3pt}	
	\setlength{\belowdisplayskip}{-3pt}
	x^k_i = W^k_x x^k_i + W^{k'}_{\alpha} \sum_{k'}{\sum_{j}{\alpha_{ij}x^{k'}_j}}
	\end{equation}
%\vspace{-0.1cm}
\end{center}
where $W^k_x$ and $W^{k'}_{\alpha}$ are learnable matrices.
The $k$ and $k'$ are node type indexes and $k' \neq k$.
The $i$ and $j$ are node indexes.
The $\alpha_{ij} = \frac{w_{ij}}{\sum_{j,j\neq i}{w_{ij}}}$ is the normalized value of the heterogeneous edge weight defined in Section~\ref{subsubsec:localgraph}.

\subsubsection{Edge Embedding Learning}
Compared with the node embeddings,
edge embeddings reveal much more structure information of the graph.
For our heterogenous graph,
edges are also in different types since they connect different types of nodes.
Considering that nodes have been embedded in a common semantic space
by the node message passing module,
we directly concatenate them and use a projection to extract the edge embedding:
\begin{center}
%\vspace{-0.5cm}
	\begin{equation}
	e_{ij} = W_e[x^k_i \oplus x^{k'}_j]
	\end{equation}
%\vspace{-0.2cm}
\end{center}
where $k$ and $k'$ are node type indexes and $\oplus$ means concatenation.
The $W_e$ is a learnable matrix.
If $k = k'$, $e_{ij}$ is the embedding of homogeneous edge, otherwise heterogeneous edge.

\subsubsection{Local Context Graph Representations}
For each local context graph,
we learn two kinds of representations to capture the short-term behavior information:
a semantic representation that reflects semantic meanings and a structural representation that catches information of the graph structure.

%Since node embeddings have been learned ,
We obtain the semantic representation of the local context graph by combining the embeddings of all types of nodes learned with the node message passing module.
Although the attribute of a node defined in Section~\ref{subsubsec:localgraph} could reflect its importance,
the semantic meaning of the node should also be considered.
Because a concept appearing few times in the concept sequence
may contain important factors for the health status prediction, we take advantage of soft-attention mechanism to determine the importance of different nodes and combine node embeddings to get the semantic representation:
\begin{center}
	\begin{equation} 
	\setlength{\abovedisplayskip}{-3pt}	
	\setlength{\belowdisplayskip}{-3pt}
	g_s = \sum_{k,i} \beta_i^k x^k_i 
	\end{equation}
\end{center}

\begin{center}
%\vspace{-0.3cm}
	\begin{equation} 
	\setlength{\abovedisplayskip}{-3pt}	
	\setlength{\belowdisplayskip}{-3pt}
	\beta_i^k = \frac{exp(q \cdot x^k_i)}{\sum_{k,i}exp(q \cdot x^k_i)} \label{eq: node attention}
	\end{equation}
%\vspace{-0.3cm}
\end{center}
% {\color{red}
where $g_s$ is the semantic representation for the local context graph, $\beta_i^k$ is the relevant importance given to each node when blending all nodes together, and $q$ is a trainable vector used as query.
The reason for using softmax in~\eqref{eq: node attention} mainly lies in two points.
(1) 
%softmax function is more easier to derive than L1 function. 
%
%The L1 function 
% {\color{red}
%is not differentiable
%since it contains absolute operation.
%
The softmax is differentiable thus can be easily integrated in the graph neural networks for end-to-end training.
(2) 
The output values of softmax function are in the range of
[0,1] with the sum of 1. 
With the softmax function, $\beta_i^k$ can be interpreted as the relevant importance given to node $x_i^k$ when blending all nodes together.
% while L1 normalization has no such effect. \\
% }

For the structural representation,
since different edges play various roles for the graph structure,
we also use attention mechanism to calculate their correlations and then combining them to get the graph structural representation.
Since the semantic representation $g_s$ contains the global information of the local context graph with semantic meanings of all nodes considered,
we treat it as a query vector to help learning more effective attentions for combining edge embeddings:
\begin{center}
	\begin{equation}
	\setlength{\abovedisplayskip}{-3pt}	
	\setlength{\belowdisplayskip}{-3pt}
	g_e = \sum_{i,j} \beta_{ij} e_{ij}
	\end{equation}
\end{center}
\begin{center}
%\vspace{-0.1cm}
 	\begin{equation}
 	\setlength{\abovedisplayskip}{-3pt}	
 	\setlength{\belowdisplayskip}{-3pt}
 	\beta_{ij} = \frac{exp[(W_{\beta}g_s) \cdot e_{ij}]}{\sum_{i,j}exp[(W_{\beta}g_s) \cdot e_{ij}]}.	\label{eq: edge attention}
 	\end{equation}
% \vspace{-0.1cm}
\end{center}
where $e_{ij}$ is the embedding for either homogeneous edges or heterogeneous edges.
$\beta_{ij}$ is the relevant importance given to each edge when blending all edges together, and $W_{\beta}$ is a projection matrix for semantic representation $g_s$.

We get the final representation for each local context graph with the concatenation of its semantic and structural representations as $g = [g_e;g_s]$.

\subsection{Global Temporal Relation Modeling}
Self-attention network (SAN) is introduced in Transformer\cite{vaswani2017attention} for the first time,
which has a sequence-to-sequence architecture and is popularly used in neural machine translation.
Taking a token sequence as input,
SAN calculates the attention scores between each token and other tokens with multiple attention heads.
Then the token embeddings are updated with other token embeddings according to their attention scores.
From this perspective,
SAN can be regarded as a graph neural network with token sequence as fully-connected nodes,
while the multi-head attention mechanism is a special message passing method.
Inspired by this,
we implement the global temporal relation modeling with the self-attention network.

Specifically,
we can get a sequence of all local context graph representations
$\{g_1, g_2, \cdots, g_T\}$ as illustrated in Section~\ref{subsec:localgraphmodel}.
For the temporal information among different local graphs,
we adopt position embeddings in \cite{vaswani2017attention} to encode the relative position of each local context graph,
noted as  $\{p_1, p_2, \cdots, p_T\}$.
Therefore, the representation for the $i$th graph is the sum of $g_i$ and $p_i$.
Then the correlations between any two local context graphs can be calculated with attention scheme:
\begin{center}
	\begin{equation}
	\setlength{\abovedisplayskip}{-5pt}	
	\setlength{\belowdisplayskip}{-3pt}
	f(g_i, g_j) = {[W_q(g_i + p_i)]}^{\top}[W_p(g_j+p_j)]
	\end{equation}
\end{center}
where $W_q$ and $W_p$ are learnable matrices.

The attention scores are scaled and normalized with a Softmax function,
which are used to get an attended representation for each local context graph $g_i$:
\begin{center}
	\begin{equation}
	\setlength{\abovedisplayskip}{-3pt}	
	\setlength{\belowdisplayskip}{-3pt}
	 \gamma_{ij} = \frac{exp(f(g_i, g_j)/\sqrt{d_p})}{\sum_{j=1}^{T}{exp(f(g_i, g_j)/\sqrt{d_p})}} \label{eq:day attention}
	\end{equation}
\end{center}
\begin{center}
	\begin{equation}
	\setlength{\abovedisplayskip}{-3pt}	
	\setlength{\belowdisplayskip}{-3pt}
	g'_i = \sum_{j=1}^{T}{\gamma_{ij}W_gg_j}
	\end{equation}
\end{center}
where $W_g$ is a learnable projection matrix,
and $d_p$ is the dimension of $g_i$.
$T$ is the number of local context graphs.
$g'_i$ is the attended representation for the $i$th local context graph.
Finally, we can get the structure-aware representation $g^*$ of the global temporal graph $\mathcal{G}$ by adding
the attended representations of all local context graphs.

\subsection{Objective Function}

The final loss function is written as the sum of a classification loss and a node variance constraint:
%
%\begin{center}
%	\begin{equation}
%	\setlength{\abovedisplayskip}{-3pt}	
%	\setlength{\belowdisplayskip}{-3pt}
$L = L_c + \lambda L_n$,
%	\end{equation}
%\end{center}
where $\lambda$ is a trade-off parameter.

\textbf{Classification Loss:}
With the representation $g^*$ of the global temporal graph,
we predict the health status label $y$ by a fully-connected layer with Softmax activation.
%\begin{center}
%	\begin{equation}
%	y = W_cg^*
%	\end{equation}
%\end{center}
%where $W_c$ is a learnable projection matrix, $G$ is the global temporal graph representation, $y$ is the predicted class.
Then we calculate the cross-entropy loss:
\begin{center}
%\vspace{-0.4cm}
	\begin{equation}
	\setlength{\abovedisplayskip}{-3pt}	
	\setlength{\belowdisplayskip}{-3pt}
	L_c = \frac{1}{N} \sum_{i=1}^NCrossEntropy(y_i, y'_i)
%	L^t = \lambda [\frac{1}{m}\sum_{i=1}^m (y_i^p-{y'}_i^p)^2] + (1-\lambda)[\frac{1}{n}\sum_{j=1}^n (y_j^g-{y'}_j^g)^2]
	\end{equation}
%\vspace{-0.2cm}
\end{center}
where $N$ is the number of instances used in training process
and $y'_i$ is the groundtruth label of the health status.
%and $\theta$ is a trade-off hyperparameter.

\textbf{Node Variance Loss:}
It is worth noting that many GNNs will face the problem of node homogenization after several epoches of node message passing since all nodes exchange information with their neighbors.
In the local context graph modeling illustrated in Section~\ref{subsec:localgraphmodel},
the message passing is not only applied on homogeneous nodes but also on heterogeneous nodes,
which may make the representations of nodes similar.
To alleviate this problem,
we add a constraint $L_n$ to the node representations to control the variance of all nodes.
Specifically, we concatenate all the node embeddings into a matrix noted as
$E \in \mathbb{R}^{N \times d}$,
where $N$ is the number of nodes of all types and
$d$ is the dimension of the node embedding.
Then we calculate the variance and get a vector $v \in \mathbb{R}^d$,
where each element represents the variance of the corresponding dimension in $E$.
Finally,
the node variance loss is defined as the average of the elements in the variance vector followed by a sigmoid function:
%
%{\color{red}
\begin{center}
%\vspace{-0.4cm}
	\begin{equation}
	\setlength{\abovedisplayskip}{-3pt}	
	\setlength{\belowdisplayskip}{-3pt}
	L_n = -sigmoid(\frac{1}{d}\sum_{i=1}^dv_i)
	%	L^t = \lambda [\frac{1}{m}\sum_{i=1}^m (y_i^p-{y'}_i^p)^2] + (1-\lambda)[\frac{1}{n}\sum_{j=1}^n (y_j^g-{y'}_j^g)^2]
	\end{equation}
%\vspace{-0.2cm}
\end{center}
%}%%%Ôö¼ÓÁË¸ººÅ£¬Çë¼ì²é

\section{Experiments}
\subsection{Dataset}\label{subsec:data}
%\subsubsection{Dataset Introduction}
We evaluate the performance of our method on StudentLife dataset~\cite{wang2014studentlife}
which is collected from 48 Dartmouth students over a 10 week term.
It contains sensor data,
EMA data (Ecological Momentary Assessment, i.e. stress),
pre and post survey responses (i.e. PHQ-9 depression scale) and educational data (GPA).
%assessing particular events in subjects' lives or assess subjects at periodic intervals
%%%% ½âÊÍEMAºÍPAM£¬
%
% Multiple sensors are used to collect the real time daily data streams,
% including accelerometer, microphone, GPS, Bluetooth, light sensor, phone charge, phone lock, and WiFi.
% %
% The EMA data, survey responses and educational data are collected as annotations that measure
% health-related states of student.
%
% {\color{red}
During the 10 weeks, students carry their phones throughout the day.
Data streams from multiple sensors, including accelerometer, microphone, GPS, Bluetooth, light sensor, phone charge, phone lock, and WiFi are 
collected in real time by the mobile phone.
Besides, students are asked to respond to various EMA questions and surveys, which are provided by psychologists to measure their mental health status.
Educational performance data, such as the grades are also collected.
In our experiment,
we use data streams collected by three representative sensors (i.e., accelerometer, microphone, WiFi) as the input, since these three kinds of sensor data
%are collected more evenly and 
contain most useful information to reflect the individual behaviors.
% }
%
The reason for collecting location information with wifi instead of GPS is
that most of students' activities are in indoor environment,
where the college's WiFi AP deployment is more effective to accurately infer the location information than the GPS.

For the groundtruth annotations of the multi-source data streams,
we use \textit{photographic affect meter (PAM)}~\cite{pollak2011pam}
values in EMA data.
%to evaluate our proposed method.
%all multi-data streams are annotated with
%
%As shown in Figure \ref{PAM},
%PAM is an application run on mobile phone with 16 photos shown to the user, and user needs to choose one photo that best describes his feeling now.
%
%Each photo corresponds to a score between 1-16 mapped to the .
%
The PAM value practically represents a score between 1-16 which
represents the Positive and Negative Affect Schedule (PANAS)\cite{watson1988development} and reflects the instantaneous mental health status of users.
The annotations are collected by a mobile application which captures user' feeling according to user' preference of specific photos.
To keep with the conceptualization of PANAS which ranges from low pleasure and low arousal to high pleasure and high arousal,
the PAM score is further divided into 4 quadrants: negative valence and low arousal with score 1-4, negative valence and high arousal with score 5-8, positive valence and low arousal with score 9-12, and positive valence and high arousal with score 13-16.
Following \cite{pollak2011pam},
we map the PAM value into the above 4 classes.
%and {\color{red}do a 4-class classification.}.

\begin{figure}[t]
	\centering
	%\vspace{-0.4cm}
	%\setlength{\abovecaptionskip}{-0.03cm}
	%\setlength{\belowcaptionskip}{-0.2cm}
	\subfigure[Sample number for each student]{
		\begin{minipage}[t]{0.5\linewidth}
			\centering
			\includegraphics[width=6cm, height=4cm]{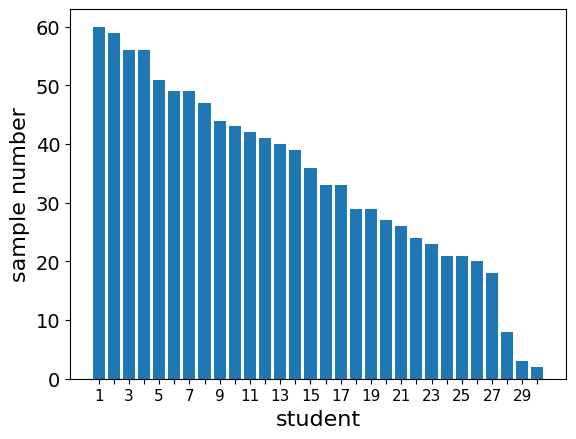}
			%\caption{fig1}
		\end{minipage}%
	}%
	\subfigure[Label distribution]{
		\begin{minipage}[t]{0.5\linewidth}
			\centering
			\includegraphics[width=6cm, height=4.5cm]{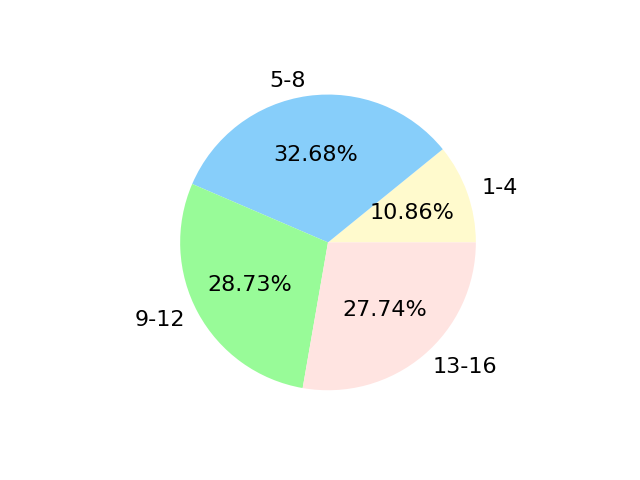}
			%\caption{fig2}
		\end{minipage}%
	}%
	\centering
	\caption{{Dataset information. (a) The number of samples for each student. (b) The distribution over 4 classes.}}
	\label{dataset}
\end{figure}

%{\color{red}
Finally,
we use data streams of 30 students who have valid PAM annotations.
%}
%%%%¼ì²éÉÏÃæÕâÖÖËµ·¨ÊÇ·ñºÏÀí
%in StudentLife to predict their daily mental health statuses.
%
Specifically, we get 912 samples in total, each sample consists of 3-day multi-modal data streams
collected by the sensors of accelerometer, microphone, and wifi.
For each sample, the instantaneous PAM label of the last day is regarded as the groundtruth label of the whole data streams. 
The sample number for each student and the sample distribution on 4 classes are shown in Figure~\ref{dataset}.
%
%This is designed with the consideration that the student's mental health status is influenced not only by the events in current day, but also the experience in last two days.
%%
%For the data split, we use 80\% of samples for training, 10\% for validation and 10\% for test.
% {\color{red}
Up to now, the mental health status prediction task in our experiment is practically a 4-class classification problem.
% }
For the training and test, we split our datasets into 10 splits and build 10 tasks on them. Each task takes 9 splits for training and the remaining one for test.
We also show the average results of all 10 tasks.
%as test we evaluate our method in a 10-fold way. Specifically, we split our dataset into 10 folds. Then we take 9 folds for training and the remaining one for test each time. The final performance is the average of test results in 10 times. Compared with using a fixed test set, we can get a more stable and convincing result in this manner.
%}
%\subsubsection{Dataset Discussion}

% {\color{red}
\textbf{Discussion on data selection.}
%In this work, we only use the StudentLife dataset to evaluate our method, since it is the only public dataset that satisfies our requirements.
% On the one hand, the data should be collected from both the healthy and unhealthy people.
% On the other hand, the sensor data we take advantage of should be continuous and long-term,
%where the \textit{continuous} means that
% the data is recorded as long as the people carry their wearable devices, such as mobile phone and smart wristband.
%Some work has been done on this task with similar datasets, but all these data does not be released. The remaining medical datasets such as MIMIC-III~\cite{johnson2016mimic} and ANDI~\cite{ridge2018assembly}, although available, do not satisfy the need of our task for the two reasons. (1) These medical datasets pay more attention to disease analysis where the studied people are mainly patients, while our work focuses on personal health and daily behaviors are considered for both healthy and unhealthy people.
%(2) These medical datasets are discretely collected from patients during the hospital treatment process with professional medical equipment, such as medical imaging, while the long-term sensor data we use is continuously collected from daily life.
In this paper, our aim is to model the personal behavior and predict the health status based on the multi-source sensor data collected from people's daily life.
To the best of our knowledge,
StudentLife is the only public dataset that satisfies our requirements.
%
%{\color{red}
On the one hand, the data should be collected from both the healthy and unhealthy people.
%}
%
On the other hand, the sensor data we take advantage of should be continuous and long-term,
i.e., 
%where the \textit{continuous} means that
the data are recorded as long as the people carry their wearable devices, such as mobile phone and smart wristband.
Some work has been done on this task. 
For example, \cite{canzian2015trajectories} uses GPS data to predict depression with a SVM classifier.
\cite{burns2011harnessing} predicts depression based on GPS, accelerometer and light sensor data from smartphone.
However, all these work does not release their data.
%So we have no choice but do our work on this dataset.\\
%
Although some medical datasets, such as MIMIC-III~\cite{johnson2016mimic} and ANDI~\cite{ridge2018assembly},
%which uses Electronic Health Record(EHR) data 
have been widely used to predict diseases~\cite{che2017rnn} or other medical related events~\cite{futoma2015comparison}, 
%they are actually different from ours.
%
%The commonly used datasets on this kind of task mainly include , etc.
%
%That's to say, 
they do not satisfy the need of our task.
Firstly, these medical datasets pay more attention to disease analysis where the studied people are mainly patients.
By contrast, our work focuses on personal health, where daily behaviors are considered for both healthy and unhealthy people.
Secondly, these medical datasets are discretely collected from patients during the hospital treatment process with professional medical equipment, such as medical imaging, while the long-term sensor data we use is continuously collected from daily life.\\
% }

\subsection{Implementation Details}\label{subsec:impl}
As introduced in Section \ref{sec:method},
to create the behavior graph which reflects the structure information
contained in the multi-source data streams,
we firstly need to detect the behavior-related concepts contained in the data streams.
Here, three backbone models proposed in~\cite{lane2011bewell} are adopted to get the middle-level semantic concepts from raw sensor data.
For the accelerometer,
a decision tree model for \textit{activity} classification is used with the features extracted from the accelerometer stream to infer the concept class (i.e. $stationary$, $walking$, $running$, and $unknown$).
For the microphone,
the $audio$ data are classified into concept classes (i.e. $silence$, $voice$, $noise$, and $other$) with a HMM model.
As for the wifi,
students' wifi scan logs are firstly recorded, then the $location$ concept classes,
such as \textit{in[dana-library]},
% and \textit{near[dana-library; lsb; vail; ]}
are inferred according to the WiFi AP deployment information,
which results in a number of 9037 classes.
The $location$ classes are in a long-tail distribution with many classes appear few times,
hence we choose the top 100 most frequent classes, which cover 93\% of the $location$ data.

After obtaining the three kinds of detected concepts, we cut the sequences in days and build the local context graph and the global temporal graph
to predict mental health status with the method illustrated in Section~\ref{sec:method}.
%According to the illustrations of the dataset in Section~\ref{subsec:data}, the mental health status prediction task in our experiment is practically a 4-class classification problem, thus 
We use the metrics of accuracy, precision, recall and F1-score to evaluate our model. 
Accuracy is the ratio of correctly predicted samples to total predicted samples. Precision, recall and F1-score are firstly calulated in each class and then weighted by the sample number of each class.

% {\color{red}
\textbf{Discussion on data synchronization.}
%Besides, an important issue for sensor data based methods is the synchronization of different sensors. 
%
In existing health related systems and methods for analyzing wearable sensor data, 
such as the risk situation recognition system~\cite{yebda2019multi}, %propose  with a focus on data synchronisation.
synchronization of different sensors is a very important issue.
Specifically, the system always contains several devices to collect different kinds of sensor data as well as a smartphone to receive the data from sensors. 
An algorithm of data synchronisation is necessary since the sensors are on different devices and there exist time differences between sensor data generated by sensors and received by smartphones.
As for our work, 
the sensors here we use are all embedded in the same smartphone~\cite{wang2014studentlife}, 
which have a common reference time naturally and do not have the time error when sending and receiving the data, thus the synchronisation is not essential.
% }
%}%%%ÈýÖÖÖ¸±êÒªÓÐÒ»¶¨µÄ½âÊÍ£¬accuracy, weighted precision, weighted recall£¬Ò»°ãµÄ·½·¨²»Ò»¶¨ÓÃÕâ¸ö

%%%%%ÉÏÃæÕâ¸öÍ¼Ã»ÓÐÔÚÕýÎÄÒýÓÃ£¬Ó¦¸ÃÌí¼ÓÒ»ÏÂ£¬½éÉÜÒ»ÏÂÕâ¸öÍ¼ÀïµÄÄÚÈÝ

\subsection{Compared Methods}\label{subsec:compare_method}
% {\color{red}
Since there are no previous work on PAM prediction on the same dataset,
we compare our method 
with three popularly used conventional machine learning algorithms (i.e., RF~\cite{598994}, KNN~\cite{Laaksonen2002Classification}, and SVM~\cite{burges1998tutorial})
and two deep learning algorithms (i.e., DNN~\cite{lecun2015deep}, LSTM~\cite{hochreiter1997long}).
To apply these baselines on multi-source data streams collected from wearable devices, 
we compute the~\textit{behavior feature}
by extracting a 108-dimensional feature vector
to represent the duration time of three kinds of concept classes in one day.
Specifically, the $activity$ concept takes 4 elements of the vector, which represent the duration time of \textit{stationary, walking, running, unknown} in one day.
The $audio$ concept takes 4 elements, which represent the duration time of \textit{silence, voice, noise, other} in one day.
The $location$ concept takes 100 elements, which represent the duration time of 100 locations student stay in one day.
%
%Then the concatenated features of three days are used to predict the PAM.
%
The~\textit{behavior feature} contains the principle information of the individual life,
such as where did she/he go in that day and how long did she/he communicate with other people.
We also compare our method with a recent GNN-based method (i.e., HAN~\cite{wang2019heterogeneous}) which can capture the structural information of a graph with different kinds of nodes.
%
%Besides, to figure out the effectiveness of each module in our framework,
%we do ablation study and compare the performances of the variants and the whole model.
%
Details of the compared baselines and the variants of our method are illustrated as follows.
% }

\begin{itemize}

	\item \textbf{RF\cite{598994}:} 
	This method uses Random Forest to do the classification. 
% 	{\color{red}
	Specifically, we concatenate the~\textit{behavior features} of three days to get a 324-dimensional feature. 
	Then we input it to the Random Forest.
% 	}
	%\item \textbf{Baseline-DT}. This method uses Decision Tree to do the classification with our extracted feature.
	\item \textbf{KNN\cite{Laaksonen2002Classification}:} 
	This method uses K-nearest neighbor.
% 	{\color{red}
	We first get the 324-dimensional feature of continuous three days as in~\textbf{RF}. 
	Then we train a K-nearest neighbor based on the feature.
% 	}
	
	\item \textbf{SVM\cite{burges1998tutorial}:}
	This method trains SVM to do the classification.
% 	{\color{red}
	Features are obtained as in~\textbf{RF}, and then a SVM is trained to predict the health status.
% 	}
	
%	{\color{red}
	\item \textbf{DNN\cite{lecun2015deep}:}
% 	{\color{red}
	This method uses
	%the basic deep neural network. We design 
	a two-layer deep neural network with the input feature computed as in~\textbf{RF}.
	Each layer is fully-connected and the hidden size is 50, which is determined by cross-validation.
% 	}
	
%	{\color{red}
	\item \textbf{LSTM\cite{hochreiter1997long}:}
	This method uses a LSTM with the hidden size of 100 to capture the temporal information of sequences.
% 	{\color{red}
    Specifically, we transform the 3-day data into a sequence with the length of 3.
    Each element in the sequence is a 108-dimensional~\textit{behavior feature}. %obtained as described above. 
    Then the sequence is input into LSTM, and the hidden state at the last step is used to predict the health status.
    % }
%}
	
	%{\color{red}
	\item \textbf{HAN\cite{wang2019heterogeneous}:}
	%
	%{\color{red}
	This method uses the heterogeneous attention network~\cite{wang2019heterogeneous} instead of our local context graph modeling method to learn the node embedding and graph representation.
% 	{\color{red}
	Specifically, we get the meta-path based neighbors for each kind of nodes according to our local context graph. 
	%}
	%
	Then the node-level attention and semantic-level attention are performed as \textbf{HAN\cite{wang2019heterogeneous}} to get the local context graph representation, which is finally input into the self-attention network to predict the health status.
% 	}

	\item \textbf{Ours$\backslash$he:}
	This variant of the proposed method omits the heterogeneous message passing in Section 3.4.1 while keeps the homogeneous message passing in the heterogeneous graph neural network.
	
	\item \textbf{Ours$\backslash$ho:}
	This variant omits the homogeneous message passing in Section 3.4.1 while keeps the heterogeneous message passing in the heterogeneous graph neural network. 
	It is used to compare with Ours$\backslash$he to illustrate the impact of the homogeneous and heterogeneous message passing.

	\item \textbf{Ours$\backslash$s:}
	In this variant, we omit the semantic representation learned by Equation (4), and only use the structural representation of the local context graph.
    % {\color{red}
	The structural representation is then input into the self-attention network to get the global temporal graph representation.
% 	}
	%

	\item \textbf{Ours$\backslash$t:}
	In this variant, we omit the structural representation learned by Equation (6), and only use the semantic representation of the local context graph.
% 	{\color{red}
	The semantic representation is then input into the self-attention network to get the global temporal graph representation.

	\item \textbf{Ours$\backslash$g:}
	In this variant, 
	representations of all local context graphs are directly added to get the final global temporal graph representation without using the self-attention network.\\
  
\end{itemize}

\renewcommand{\arraystretch}{1.5}

\begin{table}[tp]
	\setlength\tabcolsep{1pt}
	\scriptsize
	\centering
	\caption{PAM prediction results.(A: Accuracy, P: Precision, R: Recall, F1: F1 score)}\label{tab:res}
	%\setlength{\abovecaptionskip}{-0.04cm}
	%\fontsize{6.5}{8}\selectfont
	
	\begin{threeparttable}
		
		%		\label{tab:performance_comparison}
		\begin{tabular}{ccccccccccccccccccccc}
			\toprule
			\multirow{2}{*}{Method}&
			\multicolumn{4}{c}{T1}&\multicolumn{4}{c}{T2}&\multicolumn{4}{c}{T3}&\multicolumn{4}{c}{T4}&\multicolumn{4}{c}{T5} \cr
			\cmidrule{2-5} \cmidrule{6-9} \cmidrule{10-13} \cmidrule{14-17} \cmidrule{18-21}
			& A & P & R &F1 & A & P & R &F1 & A & P & R &F1 & A & P & R &F1 & A & P & R &F1  \cr
			\midrule
			RF\cite{598994}
			& 0.34  & 0.31  & 0.34  & 0.32
			& 0.33  &  0.30 & 0.33  & 0.31
			& 0.33  &  0.29 & 0.33  & 0.31
			& 0.34  &  0.30 & 0.34  & 0.32
			& 0.34  &  0.30 & 0.34  & 0.32
			%			& 0.33  &  0.34 & 0.33  & 0.33
			%			& 0.33  &  0.34 & 0.33  & 0.33
			%			& 0.33  &  0.29 & 0.33  & 0.31
			%			& 0.34  &  0.30 & 0.34  & 0.32
			%			& 0.33  &  0.29 & 0.33  & 0.31
			\cr
			KNN\cite{Laaksonen2002Classification}
			&  0.36  & 0.32 & 0.36 & 0.34
			&  0.36  & 0.31 & 0.36 & 0.34
			&  0.35  & 0.31 & 0.35 & 0.33
			&  0.34  & 0.30 & 0.34 & 0.32
			&  0.35  & 0.31 & 0.35 & 0.33
			%			&  0.35  & 0.31 & 0.35 & 0.33
			%			&  0.36  & 0.31 & 0.36 & 0.33
			%			&  0.36  & 0.34 & 0.36 & 0.35
			%			&  0.35  & 0.31 & 0.35 & 0.33
			%			&  0.35  & 0.34 & 0.35 & 0.34
			\cr
			SVM\cite{burges1998tutorial}
			&   0.37  & 0.37  & 0.37  & 0.37
			&   0.37  &  0.33 & 0.37  & 0.35
			&   0.36  &  0.32 & 0.36  & 0.34
			&   0.36  &  0.31 & 0.36  & 0.34
			&   0.37  &  0.33 & 0.37  & 0.35
			%			&   0.35  &  0.31 & 0.35  & 0.33
			%			&   0.36  &  0.31 & 0.36  & 0.34
			%			&   0.37  &  0.33 & 0.37  & 0.35
			%			&   0.37  &  0.33 & 0.37  & 0.35
			%			&   0.37  & 0.37  & 0.37  & 0.37
			\cr
			\midrule
			DNN\cite{lecun2015deep}
			&  0.39 & 0.35 & 0.39  &  0.35
			&  0.39 & 0.34 & 0.39  &  0.36
			&  0.40 & 0.34 & 0.40  &  0.37
			&  0.37 & 0.33 & 0.37  & 0.35
			&  0.39 & 0.34 & 0.39  &  0.36
			%		    &  0.40 & 0.34 & 0.40  &  0.37
			%			&  0.37 & 0.37 & 0.37  &  0.37
			%			&  0.39 & 0.34 & 0.39  &  0.36
			%			&  0.37 & 0.33 & 0.37  & 0.35
			%		    &  0.39 & 0.33 & 0.39  &  0.36
			\cr
			LSTM\cite{hochreiter1997long}
			&   0.41  & 0.37 &  0.41  &  0.39
			&   0.41  & 0.35 & 0.41   & 0.37
			&   0.42  & 0.36 &  0.42  &  0.38
			&   0.43  & 0.41 & 0.43   & 0.41
			&   0.41  & 0.35 & 0.41   & 0.37
			%			&   0.42  & 0.36 &  0.42  &  0.38
			%			&   0.41  & 0.35 & 0.41   & 0.37
			%			&   0.42  & 0.36 &  0.42  &  0.38
			%			&   0.42  & 0.40 &  0.42  &  0.41
			%			&   0.41  & 0.35 & 0.41   & 0.37
			\cr
			HAN\cite{wang2019heterogeneous}
			&   0.42  &  0.39 &  0.42 &  0.40
			&   0.43  &  0.41 & 0.43  & 0.41
			&   0.42  &  0.40 &  0.42 &  0.41
			&   0.42  &  0.41 & 0.44  & 0.42
			&   0.42  &  0.39 &  0.42 &  0.40
			%			&   0.43  &  0.41 & 0.43  & 0.41
			%			&   0.43  &  0.40 &  0.43 &  0.40
			%			&   0.43  &  0.41 & 0.43  & 0.41
			%			&   0.42  &  0.39 &  0.42 &  0.40
			%		    &   0.42  &  0.41 & 0.44  & 0.42
			\cr
			\midrule
			Ours
			&\textbf{0.47} &\textbf{0.44}&\textbf{0.47}&\textbf{0.43}
			&\textbf{0.48} &\textbf{0.42}&\textbf{0.48}&\textbf{0.44}
			&\textbf{0.47} &\textbf{0.46}&\textbf{0.47}&\textbf{0.45}
			&\textbf{0.46} &\textbf{0.45}&\textbf{0.46}&\textbf{0.44}
			&\textbf{0.48} &\textbf{0.47}&\textbf{0.48}&\textbf{0.46}
			%			&\textbf{0.47} &\textbf{0.46}&\textbf{0.47}&\textbf{0.45}
			%			&\textbf{0.47} &\textbf{0.46}&\textbf{0.47}&\textbf{0.45}
			%			&\textbf{0.47} &\textbf{0.43}&\textbf{0.47}&\textbf{0.44}
			%		    &\textbf{0.48} &\textbf{0.47}&\textbf{0.48}&\textbf{0.46}
			%			&\textbf{0.47} &\textbf{0.46}&\textbf{0.47}&\textbf{0.45}
			\cr
			\bottomrule
		\end{tabular}
	\end{threeparttable}

	\begin{threeparttable}
		%		\label{tab:performance_comparison}
		\begin{tabular}{ccccccccccccccccccccc}
			\toprule
			\multirow{2}{*}{Method}&
			\multicolumn{4}{c}{T6}&\multicolumn{4}{c}{T7}&\multicolumn{4}{c}{T8}&\multicolumn{4}{c}{T9}&\multicolumn{4}{c}{T10} \cr
			\cmidrule{2-5} \cmidrule{6-9} \cmidrule{10-13} \cmidrule{14-17} \cmidrule{18-21}
			& A & P & R &F1 & A & P & R &F1 & A & P & R &F1 & A & P & R &F1 & A & P & R & F1  \cr
			\midrule
			RF\cite{598994}
			%			& 0.34  & 0.31  & 0.34  & 0.32
			%			& 0.33  &  0.30 & 0.33  & 0.31
			%			& 0.33  &  0.29 & 0.33  & 0.31
			%			& 0.34  &  0.30 & 0.34  & 0.32
			%			& 0.34  &  0.30 & 0.34  & 0.32
			& 0.33  &  0.34 & 0.33  & 0.33
			& 0.33  &  0.34 & 0.33  & 0.33
			& 0.33  &  0.29 & 0.33  & 0.31
			& 0.34  &  0.30 & 0.34  & 0.32
			& 0.33  &  0.29 & 0.33  & 0.31 \cr
			% & 0.33  &  0.31 & 0.33  & 0.32   \cr
			KNN\cite{Laaksonen2002Classification}
			%			&  0.36  & 0.32 & 0.36 & 0.34
			%			&  0.36  & 0.31 & 0.36 & 0.34
			%			&  0.35  & 0.31 & 0.35 & 0.33
			%			&  0.34  & 0.30 & 0.34 & 0.32
			%			&  0.35  & 0.31 & 0.35 & 0.33
			&  0.35  & 0.31 & 0.35 & 0.33
			&  0.36  & 0.31 & 0.36 & 0.33
			&  0.36  & 0.34 & 0.36 & 0.35
			&  0.35  & 0.31 & 0.35 & 0.33
			&  0.35  & 0.34 & 0.35 & 0.34 \cr
			%	&  0.35  & 0.32 & 0.35 & 0.33   \cr
			SVM\cite{burges1998tutorial}
			%			&   0.37  & 0.37  & 0.37  & 0.37
			%			&   0.37  &  0.33 & 0.37  & 0.35
			%			&   0.36  &  0.32 & 0.36  & 0.34
			%			&   0.36  &  0.31 & 0.36  & 0.34
			%			&   0.37  &  0.33 & 0.37  & 0.35
			&   0.35  &  0.31 & 0.35  & 0.33
			&   0.36  &  0.31 & 0.36  & 0.34
			&   0.37  &  0.33 & 0.37  & 0.35
			&   0.37  &  0.33 & 0.37  & 0.35
			&   0.37  & 0.37  & 0.37  & 0.37 \cr
			%&   0.37  & 0.33  & 0.37  & 0.35    \cr
			\midrule
			DNN\cite{lecun2015deep}
			%			&  0.39 & 0.35 & 0.39  &  0.35
			%			&  0.39 & 0.34 & 0.39  &  0.36
			%			&  0.40 & 0.34 & 0.40  &  0.37
			%			&  0.37 & 0.33 & 0.37  & 0.35
			%			&  0.39 & 0.34 & 0.39  &  0.36
			&  0.40 & 0.34 & 0.40  &  0.37
			&  0.37 & 0.37 & 0.37  &  0.37
			&  0.39 & 0.34 & 0.39  &  0.36
			&  0.37 & 0.33 & 0.37  & 0.35
			&  0.39 & 0.33 & 0.39  &  0.36 \cr
			%&  0.39 & 0.34 & 0.39  &  0.36 \cr
			LSTM\cite{hochreiter1997long}
			%			&   0.41  & 0.37 &  0.41  &  0.39
			%			&   0.41  & 0.35 & 0.41   & 0.37
			%			&   0.42  & 0.36 &  0.42  &  0.38
			%			&   0.43  & 0.41 & 0.43   & 0.41
			%			&   0.41  & 0.35 & 0.41   & 0.37
			&   0.42  & 0.36 &  0.42  &  0.38
			&   0.41  & 0.35 & 0.41   & 0.37
			&   0.42  & 0.36 &  0.42  &  0.38
			&   0.42  & 0.40 &  0.42  &  0.41
			&   0.41  & 0.35 & 0.41   & 0.37
			%&   0.42  & 0.37 & 0.42   & 0.38
			\cr
			HAN\cite{wang2019heterogeneous}
			%			&   0.42  &  0.39 &  0.42 &  0.40
			%			&   0.43  &  0.41 & 0.43  & 0.41
			%			&   0.42  &  0.40 &  0.42 &  0.41
			%			&   0.42  &  0.41 & 0.44  & 0.42
			%			&   0.42  &  0.39 &  0.42 &  0.40
			&   0.43  &  0.41 & 0.43  & 0.41
			&   0.43  &  0.40 &  0.43 &  0.40
			&   0.43  &  0.41 & 0.43  & 0.41
			&   0.42  &  0.39 &  0.42 &  0.40
			&   0.42  &  0.41 & 0.42  & 0.42 \cr
			%&   0.42  &  0.40 & 0.42  & 0.41   \cr
			\midrule
			Ours
			%			&\textbf{0.47} &\textbf{0.44}&\textbf{0.47}&\textbf{0.43}
			%			&\textbf{0.48} &\textbf{0.42}&\textbf{0.48}&\textbf{0.44}
			%			&\textbf{0.47} &\textbf{0.46}&\textbf{0.47}&\textbf{0.45}
			%			&\textbf{0.46} &\textbf{0.45}&\textbf{0.46}&\textbf{0.44}
			%			&\textbf{0.48} &\textbf{0.47}&\textbf{0.48}&\textbf{0.46}
			&\textbf{0.47} &\textbf{0.46}&\textbf{0.47}&\textbf{0.45}
			&\textbf{0.47} &\textbf{0.46}&\textbf{0.47}&\textbf{0.45}
			&\textbf{0.47} &\textbf{0.43}&\textbf{0.47}&\textbf{0.44}
			&\textbf{0.48} &\textbf{0.47}&\textbf{0.48}&\textbf{0.46}
			&\textbf{0.47} &\textbf{0.46}&\textbf{0.47}&\textbf{0.45}
			%&\textbf{0.47} &\textbf{0.45}&\textbf{0.47}&\textbf{0.45}
			\cr
			\bottomrule
		\end{tabular}
	\end{threeparttable}
	
\end{table}

\subsection{Result Analysis}
\subsubsection{Performance Comparison}

Here we show both the results on 10 tasks in Table~\ref{tab:res} and Table~\ref{tab:ablation} and the average results in Figure~\ref{result}. 
It can be seen that our model performs better than baselines on all metrics, and most variants of the proposed method also have good results.
%{\color{red}

Compared with the traditional machine learning methods, all the deep learning based methods perform better. Baseline-LSTM has better results than Baseline-DNN since it takes the temporal information into consideration. The HAN updates the node embedding and graph representation in a meta-path way with several kinds of adjacent matrix. However, this method does not consider the direct connection between heterogeneous nodes, which ignores the semantic interaction between different kinds of nodes thus
it performs worse than our method.

\begin{table}[tp]
%\vspace{-0.2cm}
%	\setlength{\abovecaptionskip}{-0.05cm}
%\setlength{\belowcaptionskip}{-0.1cm}
%\setlength{\abovecaptionskip}{-0.2cm}
\setlength\tabcolsep{1pt}
\scriptsize
\centering
\caption{Ablation study results.(A: Accuracy, P: Precision, R: Recall, F1: F1 score)}\label{tab:ablation}
%\subtable{
%	\setlength{\abovecaptionskip}{-4cm}
	%\fontsize{6.5}{8}\selectfont
	\begin{threeparttable}
		\setlength{\abovecaptionskip}{0.cm}
%		\setlength{\belowcaptionskip}{-3cm}
		%\caption{Ablation study results.}
%		\label{tab:performance_comparison}
		\begin{tabular}{ccccccccccccccccccccc}
			\toprule
			\multirow{2}{*}{Method}&
			\multicolumn{4}{c}{T1}&\multicolumn{4}{c}{T2}&\multicolumn{4}{c}{T3}&\multicolumn{4}{c}{T4}&\multicolumn{4}{c}{T5} \cr
			\cmidrule{2-5} \cmidrule{6-9} \cmidrule{10-13} \cmidrule{14-17} \cmidrule{18-21}
			& A & P & R &F1 & A & P & R &F1 & A & P & R &F1 & A & P & R &F1 & A & P & R &F1 \cr
			\midrule
			Ours$\backslash$he
			& 0.42  &  0.37 & 0.42  & 0.39
			& 0.42  &  0.38 & 0.42  & 0.38
			& 0.40  &  0.34 & 0.40  & 0.36
			& 0.44  &  0.41 & 0.44  & 0.39
			& 0.43  &  0.42 & 0.43  & 0.39 \cr
%			& 0.41  &  0.36 & 0.41  & 0.37
%			& 0.42  &  0.40 & 0.42  & 0.38
%			& 0.40  &  0.35 & 0.40  & 0.37
%			& 0.41  &  0.36 & 0.41  & 0.37
%			& 0.42  &  0.37 & 0.42  & 0.39  \cr
			Ours$\backslash$ho
			&  0.45  & 0.42 &0.45  &0.42
			&  0.45  & 0.43 & 0.45 & 0.40
			&  0.46  & 0.43 &0.46  &0.43
			&  0.45  & 0.42 & 0.45 & 0.42
			&  0.44  & 0.41 &0.44  &0.41   \cr
%			&  0.44  & 0.41 & 0.44 & 0.42
%			&  0.46  & 0.44 &0.46  &0.44
%			&  0.43  & 0.42 & 0.45 & 0.42
%			&  0.45  & 0.42 &0.45  &0.43
%			&  0.45  & 0.43 & 0.45 & 0.44   \cr
			Ours$\backslash$s
			&   0.43  &  0.40 & 0.43  & 0.39
			&   0.42  &  0.40 & 0.42  & 0.40
			&   0.43  &  0.40 & 0.43  & 0.41
			&   0.43  &  0.43 & 0.43  & 0.42
			&   0.44  &  0.41 & 0.44  & 0.42  \cr
%			&   0.43  &  0.40 & 0.43  & 0.39
%			&   0.43  &  0.44 & 0.43  & 0.42
%			&   0.42  &  \textbf{0.43} & 0.43  & 0.41
%			&   0.44  &  0.42 & 0.44  & 0.43
%			&   0.41  &  0.42 & 0.41  & 0.41  \cr
			Ours$\backslash$t
			 &  0.46 & 0.44 & 0.46  &  0.45
			 &  0.45 & 0.43 & 0.46  &  0.43
			 &  0.46 & 0.43 & 0.45  &  0.44
			 &  0.45 & 0.44 & 0.45  &  \textbf{0.44}
			 &  0.46 & 0.44 & 0.46  &  0.45 \cr
%			 &  0.44 & 0.41 & 0.44  &  0.42
%			 &  0.44 & 0.42 & 0.44  &  0.43
%			 &  0.45 & 0.42 & 0.45  &  0.42
%			 &  0.46 & 0.43 & 0.46  &  0.43
%			 &  0.46 & 0.44 & 0.46  &  0.44 \cr
			Ours$\backslash$g
			&   0.44  & 0.41 &  0.44  &  0.40
			&  0.43  & \textbf{0.44} & 0.43  & 0.42
			&   0.42  & 0.43 &  0.42  &  0.43
			&  0.44  &  0.41 & 0.44   & 0.40
			&   0.44  & 0.42 &  0.44  &  0.43 \cr
%			&  0.44  &  0.42 & 0.44   & 0.43
%			&   0.45  & 0.42 &  0.45  &  0.42
%			&  0.44  &  0.41 & 0.44   & 0.40
%			&   0.44  & 0.42 &  0.44  &  0.43
%			&  0.44  & 0.41 & 0.44   & 0.40  \cr
			Ours
			&\textbf{0.47} &\textbf{0.44}&\textbf{0.47}&\textbf{0.43}
			&\textbf{0.48} &0.42         &\textbf{0.48}&\textbf{0.44}
			&\textbf{0.47} &\textbf{0.46}&\textbf{0.47}&\textbf{0.45}
			&\textbf{0.46} &\textbf{0.45}&\textbf{0.46}&\textbf{0.44}
			&\textbf{0.48} &\textbf{0.47}&\textbf{0.48}&\textbf{0.46}
%			&\textbf{0.47} &\textbf{0.46}&\textbf{0.47}&\textbf{0.45}
%			&\textbf{0.47} &\textbf{0.46}&\textbf{0.47}&\textbf{0.45}
%			&\textbf{0.47} &\textbf{0.43}&\textbf{0.47}&\textbf{0.44}
%			&\textbf{0.48} &\textbf{0.47}&\textbf{0.48}&\textbf{0.46}
%			&\textbf{0.47} &\textbf{0.46}&\textbf{0.47}&\textbf{0.45}
		  \cr
			\bottomrule
		\end{tabular}
	\begin{tabular}{ccccccccccccccccccccc}
	\toprule
	\multirow{2}{*}{Method}&
	\multicolumn{4}{c}{T6}&\multicolumn{4}{c}{T7}&\multicolumn{4}{c}{T8}&\multicolumn{4}{c}{T9}&\multicolumn{4}{c}{T10} \cr
	\cmidrule{2-5} \cmidrule{6-9} \cmidrule{10-13} \cmidrule{14-17} \cmidrule{18-21} %\cmidrule{22-25}
	& A & P & R &F1 & A & P & R &F1 & A & P & R &F1 & A & P & R &F1 & A & P & R &F1  \cr
	\midrule
	Ours$\backslash$he
	%			& 0.42  &  0.37 & 0.42  & 0.39
	%			& 0.42  &  0.38 & 0.42  & 0.38
	%			& 0.40  &  0.34 & 0.40  & 0.36
	%			& 0.44  &  0.41 & 0.44  & 0.39
	%			& 0.43  &  0.42 & 0.43  & 0.39
	& 0.41  &  0.36 & 0.41  & 0.37
	& 0.42  &  0.40 & 0.42  & 0.38
	& 0.40  &  0.35 & 0.40  & 0.37
	& 0.41  &  0.36 & 0.41  & 0.37
	& 0.42  &  0.37 & 0.42  & 0.39
	%& 0.42  &  0.38 & 0.42  & 0.38
	\cr
	Ours$\backslash$ho
	%			&  0.45  & 0.42 &0.45 &0.42
	%			&  0.45  & 0.43 & 0.45 & 0.40
	%			&  0.46  & 0.43 &0.46 &0.43
	%			&  0.45  & 0.42 & 0.45 & 0.42
	%			&  0.44  & 0.41 &0.44 &0.41
	&  0.44  & 0.41 & 0.44 & 0.42
	&  0.46  & 0.44 &0.46 &0.44
	&  0.43  & 0.42 & 0.43 & 0.42
	&  0.45  & 0.42 &0.45 &0.43
	&  0.45  & 0.43 & 0.45 & 0.44 \cr
%	&  0.45  & 0.42 & 0.45 & 0.42   \cr
	Ours$\backslash$s
	%			&   0.43  &  0.40 & 0.43  & 0.39
	%			&   0.42  &  0.40 & 0.42  & 0.40
	%			&   0.43  &  0.40 & 0.43  & 0.41
	%			&   0.43  &  0.43 & 0.43  & 0.42
	%			&   0.44  &  0.41 & 0.44  & 0.42
	&   0.43  &  0.40 & 0.43  & 0.39
	&   0.43  &  0.44 & 0.43  & 0.42
	&   0.42  &  \textbf{0.43} & 0.42  & 0.41
	&   0.44  &  0.42 & 0.44  & 0.43
	&   0.41  &  0.42 & 0.41  & 0.41
%	&   0.43  &  0.42 & 0.43  & 0.41
	\cr
	Ours$\backslash$t
	%			&  0.46 & 0.44 & 0.46  &  0.45
	%			&  0.45 & 0.43 & 0.46  &  0.43
	%			&  0.46 & 0.43 & 0.45  &  0.44
	%			&  0.45 & 0.44 & 0.45  &  \textbf{0.44}
	%			&  0.46 & 0.44 & 0.46  &  0.45
	&  0.44 & 0.41 & 0.44  &  0.42
	&  0.44 & 0.42 & 0.44  &  0.43
	&  0.45 & 0.42 & 0.45  &  0.42
	&  0.46 & 0.43 & 0.46  &  0.43
	&  0.46 & 0.44 & 0.46  &  0.44 \cr
%	&  0.45 & 0.43 & 0.45  &  0.44 \cr
	Ours$\backslash$g
	%			&   0.44  & 0.41 &  0.44  &  0.40
	%			&  0.43  & \textbf{0.44} & 0.43  & 0.42
	%			&   0.42  & 0.43 &  0.42  &  0.43
	%			&  0.44  &  0.41 & 0.44  & 0.40
	%			&   0.44  & 0.42 &  0.44  &  0.43
	&  0.44  &  0.42 & 0.44  & 0.43
	&   0.45  & 0.42 &  0.45  &  0.42
	&  0.44  &  0.41 & 0.44  & 0.40
	&   0.44  & 0.42 &  0.44  &  0.43
	&  0.44  & 0.41 & 0.44  & 0.40
%	&  0.44  & 0.42 & 0.44  & 0.42
	\cr
	Ours
	%			&\textbf{0.47} &\textbf{0.44}&\textbf{0.47}&\textbf{0.43}
	%			&\textbf{0.48} &0.42         &\textbf{0.48}&\textbf{0.44}
	%			&\textbf{0.47} &\textbf{0.46}&\textbf{0.47}&\textbf{0.45}
	%			&\textbf{0.46} &\textbf{0.45}&\textbf{0.46}&\textbf{0.44}
	%			&\textbf{0.48} &\textbf{0.47}&\textbf{0.48}&\textbf{0.46}
	&\textbf{0.47} &\textbf{0.46}&\textbf{0.47}&\textbf{0.45}
	&\textbf{0.47} &\textbf{0.46}&\textbf{0.47}&\textbf{0.45}
	&\textbf{0.47} &\textbf{0.43}&\textbf{0.47}&\textbf{0.44}
	&\textbf{0.48} &\textbf{0.47}&\textbf{0.48}&\textbf{0.46}
	&\textbf{0.47} &\textbf{0.46}&\textbf{0.47}&\textbf{0.45}
%	&\textbf{0.47} &\textbf{0.45}&\textbf{0.47}&\textbf{0.45}
	\cr
	\bottomrule
	\end{tabular}
	\end{threeparttable}
%}
\end{table}
As for the ablation study, it can be concluded that each module in our framework plays a significant role in the performance improvement.
By comparing the full model and Ours$\backslash$he as well as Ours$\backslash$ho,
we note that the full model performs better than both two variants,
which proves the assumption that the message passing module has positive effects on the node embedding learning, and the homogeneous edges and heterogeneous edges succeed in building the homogeneous and heterogeneous node structures.
When comparing the performances of the homogeneous message passing Ours$\backslash$he
and the heterogeneous message passing Ours$\backslash$ho,
it can be seen that the heterogeneous message passing performs better than the homogeneous message passing which may be because to heterogeneous edges help learn more comprehensive embeddings by getting additional information from other types of nodes.

\begin{figure}[htbp]
	\centering
	\vspace{-0.3cm}
	\setlength{\abovecaptionskip}{-0.05cm}
	\setlength{\belowcaptionskip}{-0.1cm}
	\subfigure[Accuracy]{
		\begin{minipage}[t]{0.5\linewidth}
			\centering
			\includegraphics[width=5.5cm, height=4cm]{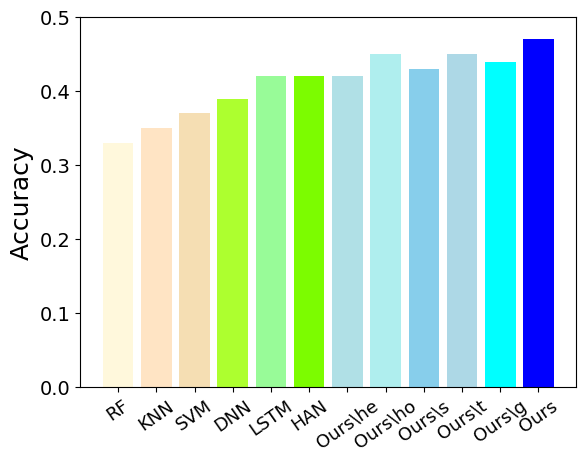}
			%\caption{fig1}
		\end{minipage}%
	}%
	\subfigure[Precision]{
		\begin{minipage}[t]{0.5\linewidth}
			\centering
			\includegraphics[width=5.5cm, height=4cm]{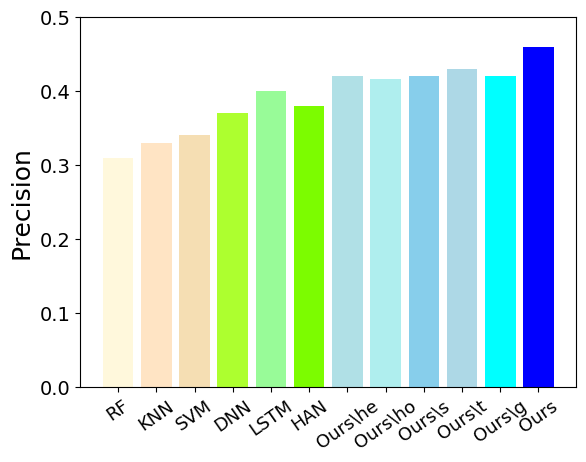}
			%\caption{fig2}
		\end{minipage}%
	}
	\subfigure[Recall]{
		\begin{minipage}[t]{0.5\linewidth}
			\centering
			\includegraphics[width=5.5cm, height=4cm]{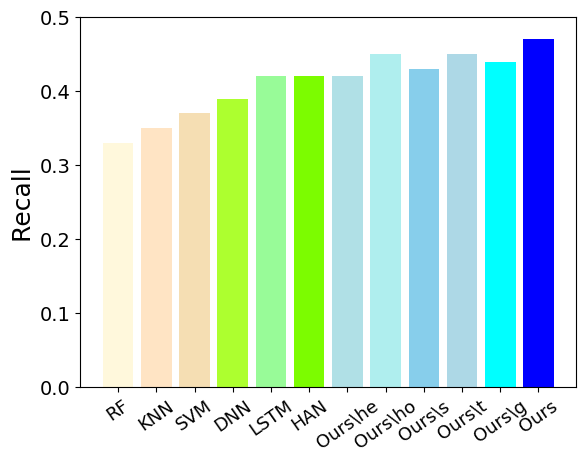}
			%\caption{fig1}
		\end{minipage}%
	}%
	\subfigure[F1-score]{
		\begin{minipage}[t]{0.5\linewidth}
			\centering
			\includegraphics[width=5.5cm, height=4cm]{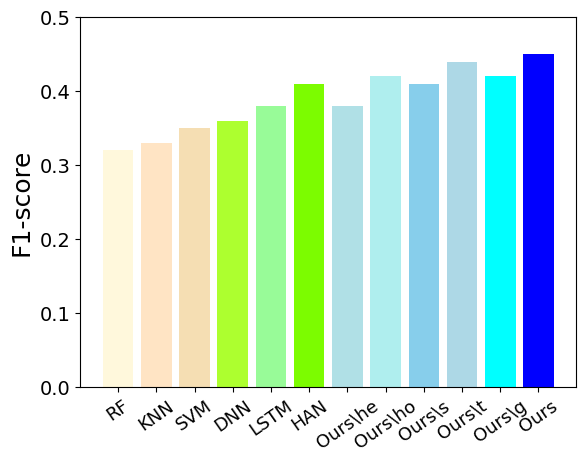}
			%\caption{fig2}
		\end{minipage}%
	}%
	\centering
	\caption{The average results on the 10 tasks.}
	\label{result}
\end{figure}

As for the comparison between Ours$\backslash$t and Ours$\backslash$s,
which use semantic representations and structural representations, respectively,
it can be noted that both the semantic representation and the structural representation benefit the prediction process,
meaning that these two kinds of representations reveal different aspects of the graph information.
The semantic representations perform better than structural representations,
which may be because the semantic representation provides more global information of the graph.
When we omit the global temporal graph while only use the representation of the local context graph to predict PAM in Ours$\backslash$g,
the performance drops a little,
meaning that not only the local context information has an positive effect on individual's health status,
but also the long-term temporal structure of behaviors makes a difference.

% {\color{red}
In Table~\ref{tab:res}, Table~\ref{tab:ablation} and Figure~\ref{result}, it is worth noting that the results of all methods are below 50\%,
which demonstrates that it is a extremely challenging task to predict mental health status based on personal behaviors in daily life, especially with limited samples.
However, the average accuracy improvement of our method accounts for about $5\%$ of the result 
%compared with
of the second best method HAN as shown in Figure~\ref{result}(a),
which still shows the advantage of the proposed method.
We believe that our model will get better performances on larger-scale datasets.
% }

%since the strucutral representations are extracted with edge embeddings, which are more effective to capture the local-global structure in multimodal data streams of the behavior.

\subsubsection{Parameter Analysis}
\begin{figure}[htb]
	\centering
	%\vspace{-0.3cm}
	\setlength{\abovecaptionskip}{-0.05cm}
	\subfigure[Node message passing module layer $m$]{
		\begin{minipage}[t]{0.45\linewidth}
			\centering
			\includegraphics[width=4cm, height=3.5cm]{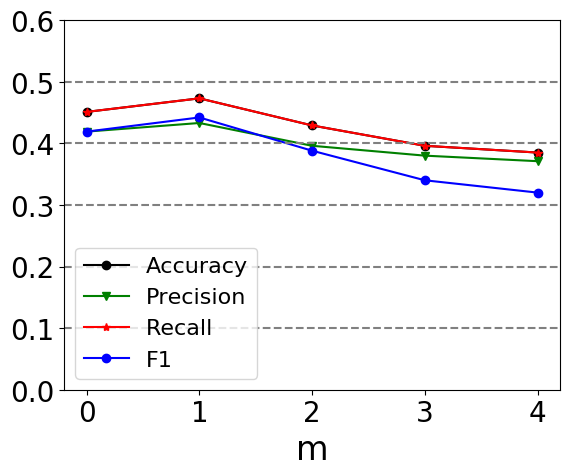}
			%\caption{fig1}
		\end{minipage}%
	}\hspace{10mm}
	\subfigure[Local context graph representation dimension $d_p$]{
		\begin{minipage}[t]{0.45\linewidth}
			\centering
	\includegraphics[width=4cm, height=3.5cm]{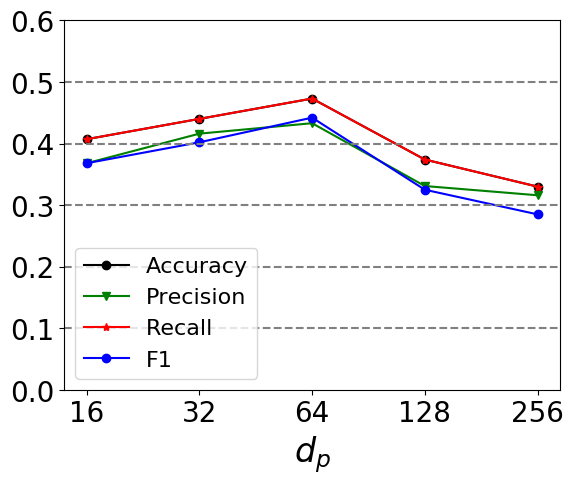}
			%\caption{fig2}
		\end{minipage}%
	}%
	\centering
	\caption{Parameter analysis of $m$, $d_p$}
	\label{parameter}
\end{figure}

Here we investigate the influence of two import hyper-parameters $m$
and $d_p$ which represent the number of layers in the local context graph
and the dimension of the final local context graph representation, respectively.
%%%%final local context graph representationµÄÎ¬¶È¾ÍÊÇnode embeddingµÄÎ¬¶È°É
We vary $m$ from 1 to 5 and keep other settings fixed,
the results on PAM prediction are shown in Figure~\ref{parameter}.
It can be noted that the performance is improved firstly with the increase of $m$.
However, when the layer number keeps increasing, the performance drops, because too many layers could make the node embedding less discriminative.
As for $d_p$, we vary it from 16 to 256 while keep other settings fixed.
The results are shown in Figure~\ref{parameter}.
We can see that the best performance is achieved in 64,
since low dimension is hard to get useful information,
while the high dimension is difficult to train with limited instances.

\begin{figure}[t]
	%\vspace{-0.2cm}
	%\setlength{\abovecaptionskip}{-0.02cm}
	%\setlength{\belowcaptionskip}{-0.2cm}
	\centering
			\includegraphics[width=12cm, height=9cm]{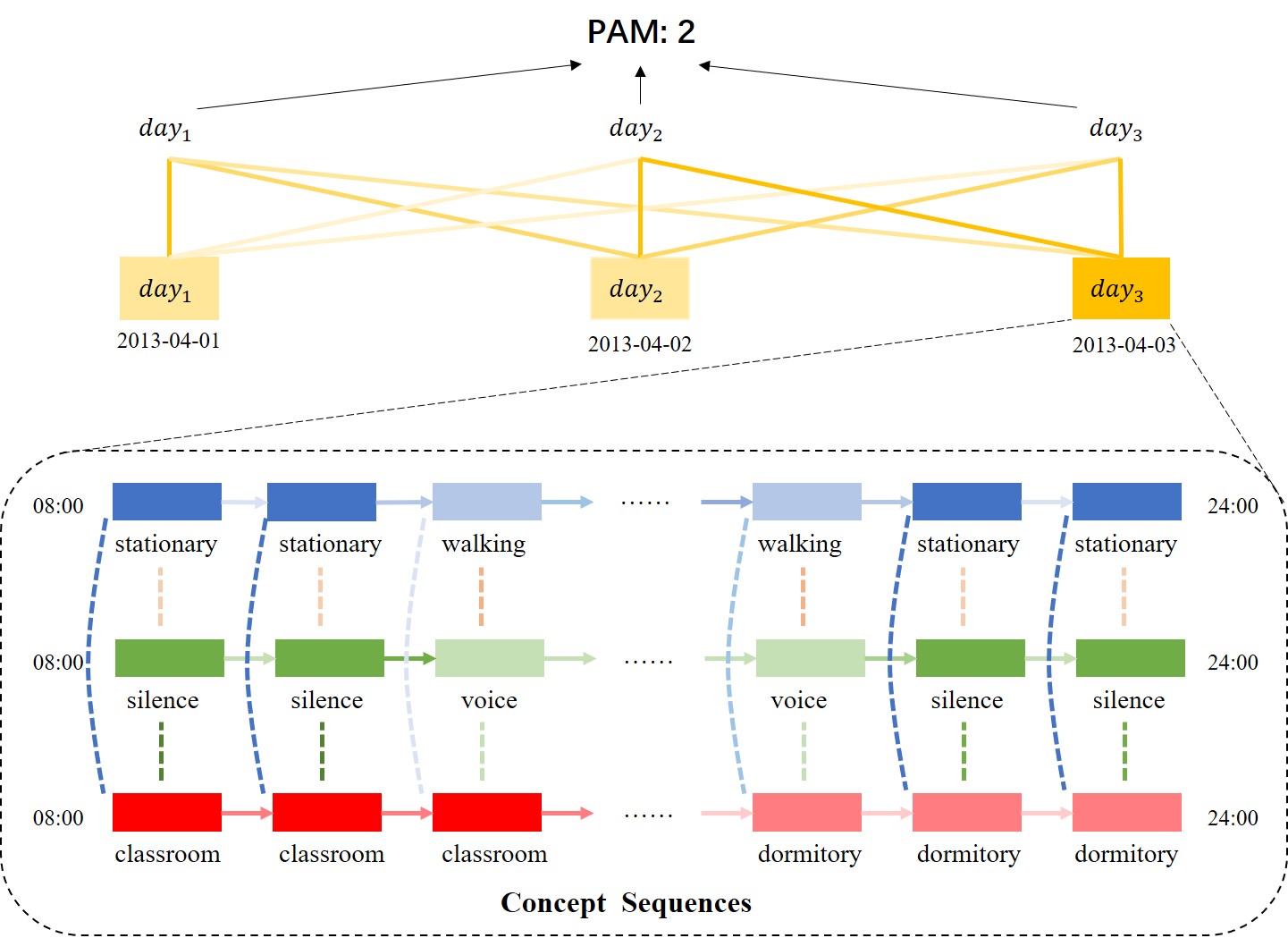}
	\centering
	\caption{Visualization of local context graphs and the global temporal graph.}
	\label{visual}
\end{figure}

\subsubsection{Visualization}
%}

Here we analyse the attention in learning representations from local context graphs and the global temporal graph to figure out the factors that influence the mental health.
Figure~\ref{visual} shows the example of an individual's 3-day data streams.
The PAM is predicted with the global representation which is learned based on representations of three local context graphs.

%We visualize the attention score in self-attention network in \ref{fig:visual}.
The attention score $\gamma_{ij}$ between two different local context graphs is computed with Eq.~\eqref{eq:day attention}.
%}
%{\color{red}
We show the attention score between any two local graphs by the darkness of the corresponding line.
The attention of each graph is represented by the darkness of the text box.
%}
%
We can see that for each local context graph,
the attention to itself tends to play the major role,
though each one pays attention to all other ones.
Besides, it can be seen that the third local context graph
has much influence on all three ones.
%
%For the third local context graph, we further visualize the concept sequences node attention score $\beta_i^k$ computed with Eq.~\ref{eq: node attention}, and the edge attention score $\beta_{ij}$ computed with Eq.~\ref{eq: edge attention}.

%{\color{red}
We further visualize the concept sequences in the third day, which are used to build the local context graph in that day. In the concept sequences, each text box represents a concept and different concept sequences are shown in different colors.
The importance of each concept is calculated with the attention mechanism in Eq.~\eqref{eq: node attention}, and represented by the darkness of text box.
%}
%
It is noted that concepts of $stationary$, $silence$ and $classroom$ play the major role in learning the representation from the local context graph.
%and all the concepts from the $location$ type receive more attention compared with the concepts from the $activity$ and $audio$ types.
%
%{\color{red}
Homogeneous edges that represent the behavior transfers are shown as arrows, while heterogeneous edges that represent behavior co-occurrences are shown as dotted lines.
The importance of the edge is also indicated by the darkness.
%}
%It is noted that homogeneous edges $stationary$-$walking$ and heterogeneous edges $classroom$-$stationary$, $classroom$-$silence$ are more important to learn the graph structural information.
%We select three homogeneous edges with the highest attention scores in three types of homogeneous edges (i.e. $activity$-$activity$, $audio$-$audio$, $location$-$location$) respectively,
%and three heterogeneous edges with the highest scores in three types of heterogeneous edges (i.e. $activity$-$location$, $activity$-$audio$, $location$-$audio$).
%
We notice that in general,
heterogeneous edges receive more attention than homogeneous edges although their occurrences are few,
which demonstrates the importance of combination of different concepts and validates that our attention mechanism successfully finds useful patterns from the multi-source data streams.
Besides, the edges, which have nodes from the $location$ type,
tend to get more attention than the ones which have nodes from the $activity$ and $audio$ types, because locations reveal much information of the daily lives.

\subsection{Application on Grade Prediction}
%{\color{red}
To better illustrate the effectiveness of our model for
learning representations from behavior-related data streams,
we propose to apply our model on the grade prediction task.
%}
%%%%%%ÉÏÃæµÄÕâ¸öÒýÎÄÀïÃæÒ²ÓÐÔ¤²âgradeµÄ·½·¨Âð£¬ÓÐµÄ»°ÕâÀïÁÐµÄbaselineÒ»ÑùÂð£¬Ò»ÑùµÄ»°£¬ÔÚtableÀï¼Ócite²»È»»áÓÐÎÊÌâ£¬
%In \cite{wang2015smartgpa}, they extract features by hand to figure out the correlation between daily behaviors and the academic performance. The features contain the student's study behaviors and social behaviors, as well as the behavior slopes which indicate the behavior changes. Then a linear regression model is used to predict grades.
%Since the sample number for each student is different, we do sampling to each student' samples with the mean sample number.
%%%%We first split the data into 30 samples according to students they belong to.
%data splitÔÚÊý¾Ý¼¯½éÉÜ¹ýÁË£¬ÕâÀïÉ¾µô£»
%{\color{red}
%Add introduction of the grade annotations.XXX.(Can also be introduced in the section of Dataset).
%}
The grade annotation is the GPA which indicates a student's overall long term academic performance in a range 0-4.

Firstly,
we take our model, which is well trained on the health status prediction task,
to extract the global representation for each student's data stream and
use KNN to do the grade prediction.
%
%To validate the effectiveness of our learned local context graph representation,
Then we compare our model with a baseline, which takes
the 108-dimensional feature introduced in Section~\ref{subsec:compare_method}
as the representation for each day and add them to get the final feature.
The same KNN is used to do the grade prediction.

We use the mean absolute errors (MAE),
the coefficient of determination ($R^2$) and Pierson correlation as evaluation metrics for the grade prediction.
We adopt the leave-one-out way to evaluate the performance.
%%%%%Èç¹ûÓÃÁôÒ»·¨µÄ»°£¬ÄÇÓÐ¶à´ÎÊµÑé½á¹û£¬ËùÒÔ±¨µÄ½á¹ûÓ¦¸Ã¶à´ÎÊµÑéµÄÆ½¾ù°É£¿ÕâÀï¸Ä³Éaverage resultsÁË£¬
The average results are shown in Table~\ref{grade}.

As shown,
our model outperforms the baseline on three metrics,
which demonstrates that our model is effective in learning the structure-aware representation of the individual's long-term behavior.
%
%{\color{red}
Moreover, our model has good generalization ability and can be used to extract global
behavior-related features in different tasks without model finetuning.
%}
%do some prediction according to the daily lives.
%as well as the self-attention network to get the final graph representations. Finally, we Since our sample number is small, we choose some traditional nonparametric methods to do the prediction.

%Comparing the 30th day and the 31th day, we can find that acitvity time ratio and audio time ratio are almost equal, the differences maily lie on location time. The 30th day spend the most time on 'classroom', which correlates more with 'study' behavior, while for the 31th day 'classroom' is much less and time on 'gymnasium' and 'club' are increased. The structural representation is a combination of all edges with different attention scores, used for both grade prediction and mental health prediction. For the edge attention scores, we notice that in general, heterogeneous edges receives more attention than horogeneous edges although their occurences are few, which demonstrates the importance of combination of different nodes and validates that our attention mechasim successfully find useful patterns in lives. Besides, the location related edges tend to get more attention than activity and audio, which may because location reveal much information of the daily lives.

\begin{table}[t]
	\centering
	\setlength{\abovecaptionskip}{-0.04cm}
	\fontsize{6.5}{8}\selectfont
	\begin{threeparttable}
		\caption{Grade prediction results.}
		\label{grade}
		\begin{tabular}{cccccc}
			\toprule
			\multirow{2}{*}{Method}& \multicolumn{3}{c}{Grade Prediction}\cr
			&MAE&$R^2$&Pearson\\
			\midrule
			Hand-crafted feature + KNN   &   0.296  &  0.10 & 0.32  \cr
			Graph representation + KNN   &  \textbf{0.195}  & \textbf{0.21} &  \textbf{0.51}  \cr
			\bottomrule
		\end{tabular}
	\end{threeparttable}
\end{table}

\section{Conclusion}
% {\color{red}
In this paper, we propose a local-global graph to model personal behavior and predict daily mental health status based on the multi-source wearable sensor data.
The graph contains multiple local context sub-graphs and a global temporal sub-graph to capture the short-term context information and long-term temporal
dependencies of individual behaviors respectively.
We learn the semantic representation and structural representation for the local context graph with heterogeneous neural network.
A self-attention network is designed to learn the representation for the global temporal
graph, which is finally used to predict the health status.
We perform experiments on the public dataset StudentLife and compare our method with popularly used machine learning and deep learning methods.
Our method outperforms all existing methods, which validates its effectiveness.
%
%However, there is still spacious room for development on this task.
In future work, we will integrate more kinds of data streams to improve the local-global individual graph, and try to apply our method on larger-scale multi-source sensor datasets for health prediction.

\bibliographystyle{ACM-Reference-Format}
\bibliography{sample-base}

%%% -*-BibTeX-*-
%%% Do NOT edit. File created by BibTeX with style
%%% ACM-Reference-Format-Journals [18-Jan-2012].

\begin{thebibliography}{72}

%%% ====================================================================
%%% NOTE TO THE USER: you can override these defaults by providing
%%% customized versions of any of these macros before the \bibliography
%%% command.  Each of them MUST provide its own final punctuation,
%%% except for \shownote{}, \showDOI{}, and \showURL{}.  The latter two
%%% do not use final punctuation, in order to avoid confusing it with
%%% the Web address.
%%%
%%% To suppress output of a particular field, define its macro to expand
%%% to an empty string, or better, \unskip, like this:
%%%
%%% \newcommand{\showDOI}[1]{\unskip}   % LaTeX syntax
%%%
%%% \def \showDOI #1{\unskip}           % plain TeX syntax
%%%
%%% ====================================================================

\ifx \showCODEN    \undefined \def \showCODEN     #1{\unskip}     \fi
\ifx \showDOI      \undefined \def \showDOI       #1{#1}\fi
\ifx \showISBNx    \undefined \def \showISBNx     #1{\unskip}     \fi
\ifx \showISBNxiii \undefined \def \showISBNxiii  #1{\unskip}     \fi
\ifx \showISSN     \undefined \def \showISSN      #1{\unskip}     \fi
\ifx \showLCCN     \undefined \def \showLCCN      #1{\unskip}     \fi
\ifx \shownote     \undefined \def \shownote      #1{#1}          \fi
\ifx \showarticletitle \undefined \def \showarticletitle #1{#1}   \fi
\ifx \showURL      \undefined \def \showURL       {\relax}        \fi
% The following commands are used for tagged output and should be
% invisible to TeX
\providecommand\bibfield[2]{#2}
\providecommand\bibinfo[2]{#2}
\providecommand\natexlab[1]{#1}
\providecommand\showeprint[2][]{arXiv:#2}

\bibitem[\protect\citeauthoryear{??}{tim}{2009}]%
        {timmurphy}
 \bibinfo{year}{2009}\natexlab{}.
\newblock
  \bibinfo{howpublished}{\url{https://www.who.int/mediacentre/multimedia/podcasts/2009/lifestyle-interventions-20090109/en/}}.
\newblock


\bibitem[\protect\citeauthoryear{??}{liv}{2017}]%
        {live}
 \bibinfo{year}{2017}\natexlab{}.
\newblock
  \bibinfo{howpublished}{\url{http://www.sdwsnews.com.cn/a/jiankangtuku/2017/1108/15306.html}}.
\newblock


\bibitem[\protect\citeauthoryear{??}{eat}{2020}]%
        {eating}
 \bibinfo{year}{2020}\natexlab{}.
\newblock \bibinfo{howpublished}{\url{https://www.who.int/}}.
\newblock


\bibitem[\protect\citeauthoryear{Acharya, Oh, Hagiwara, Tan, and Adeli}{Acharya
  et~al\mbox{.}}{2018}]%
        {acharya2018deep}
\bibfield{author}{\bibinfo{person}{U~Rajendra Acharya},
  \bibinfo{person}{Shu~Lih Oh}, \bibinfo{person}{Yuki Hagiwara},
  \bibinfo{person}{Jen~Hong Tan}, {and} \bibinfo{person}{Hojjat Adeli}.}
  \bibinfo{year}{2018}\natexlab{}.
\newblock \showarticletitle{Deep convolutional neural network for the automated
  detection and diagnosis of seizure using EEG signals}.
\newblock \bibinfo{journal}{\emph{Computers in biology and medicine}}
  \bibinfo{volume}{100} (\bibinfo{year}{2018}), \bibinfo{pages}{270--278}.
\newblock


\bibitem[\protect\citeauthoryear{Asselbergs, Ruwaard, Ejdys, Schrader,
  Sijbrandij, and Riper}{Asselbergs et~al\mbox{.}}{2016}]%
        {asselbergs2016mobile}
\bibfield{author}{\bibinfo{person}{Joost Asselbergs}, \bibinfo{person}{Jeroen
  Ruwaard}, \bibinfo{person}{Michal Ejdys}, \bibinfo{person}{Niels Schrader},
  \bibinfo{person}{Marit Sijbrandij}, {and} \bibinfo{person}{Heleen Riper}.}
  \bibinfo{year}{2016}\natexlab{}.
\newblock \showarticletitle{Mobile phone-based unobtrusive ecological momentary
  assessment of day-to-day mood: an explorative study}.
\newblock \bibinfo{journal}{\emph{Journal of medical Internet research}}
  \bibinfo{volume}{18}, \bibinfo{number}{3} (\bibinfo{year}{2016}),
  \bibinfo{pages}{e72}.
\newblock


\bibitem[\protect\citeauthoryear{Atwood and Towsley}{Atwood and
  Towsley}{2016}]%
        {atwood2016diffusion}
\bibfield{author}{\bibinfo{person}{James Atwood} {and} \bibinfo{person}{Don
  Towsley}.} \bibinfo{year}{2016}\natexlab{}.
\newblock \showarticletitle{Diffusion-convolutional neural networks}. In
  \bibinfo{booktitle}{\emph{Advances in neural information processing
  systems}}. \bibinfo{pages}{1993--2001}.
\newblock


\bibitem[\protect\citeauthoryear{B{\'a}nhalmi, Borb{\'a}s, Fidrich, Bilicki,
  Gingl, and Rudas}{B{\'a}nhalmi et~al\mbox{.}}{2018}]%
        {banhalmi2018analysis}
\bibfield{author}{\bibinfo{person}{Andr{\'a}s B{\'a}nhalmi},
  \bibinfo{person}{J{\'a}nos Borb{\'a}s}, \bibinfo{person}{M{\'a}rta Fidrich},
  \bibinfo{person}{Vilmos Bilicki}, \bibinfo{person}{Zolt{\'a}n Gingl}, {and}
  \bibinfo{person}{L{\'a}szl{\'o} Rudas}.} \bibinfo{year}{2018}\natexlab{}.
\newblock \showarticletitle{Analysis of a pulse rate variability measurement
  using a smartphone camera}.
\newblock \bibinfo{journal}{\emph{Journal of healthcare engineering}}
  \bibinfo{volume}{2018} (\bibinfo{year}{2018}).
\newblock


\bibitem[\protect\citeauthoryear{Burges}{Burges}{1998}]%
        {burges1998tutorial}
\bibfield{author}{\bibinfo{person}{Christopher~JC Burges}.}
  \bibinfo{year}{1998}\natexlab{}.
\newblock \showarticletitle{A tutorial on support vector machines for pattern
  recognition}.
\newblock \bibinfo{journal}{\emph{Data mining and knowledge discovery}}
  \bibinfo{volume}{2}, \bibinfo{number}{2} (\bibinfo{year}{1998}),
  \bibinfo{pages}{121--167}.
\newblock


\bibitem[\protect\citeauthoryear{Burns, Begale, Duffecy, Gergle, Karr,
  Giangrande, and Mohr}{Burns et~al\mbox{.}}{2011}]%
        {burns2011harnessing}
\bibfield{author}{\bibinfo{person}{Michelle~Nicole Burns},
  \bibinfo{person}{Mark Begale}, \bibinfo{person}{Jennifer Duffecy},
  \bibinfo{person}{Darren Gergle}, \bibinfo{person}{Chris~J Karr},
  \bibinfo{person}{Emily Giangrande}, {and} \bibinfo{person}{David~C Mohr}.}
  \bibinfo{year}{2011}\natexlab{}.
\newblock \showarticletitle{Harnessing context sensing to develop a mobile
  intervention for depression}.
\newblock \bibinfo{journal}{\emph{Journal of medical Internet research}}
  \bibinfo{volume}{13}, \bibinfo{number}{3} (\bibinfo{year}{2011}),
  \bibinfo{pages}{e55}.
\newblock


\bibitem[\protect\citeauthoryear{Cai, Han, Chen, Sha, Wang, Hu, Yang, Feng,
  Ding, Chen, et~al\mbox{.}}{Cai et~al\mbox{.}}{2018}]%
        {cai2018pervasive}
\bibfield{author}{\bibinfo{person}{Hanshu Cai}, \bibinfo{person}{Jiashuo Han},
  \bibinfo{person}{Yunfei Chen}, \bibinfo{person}{Xiaocong Sha},
  \bibinfo{person}{Ziyang Wang}, \bibinfo{person}{Bin Hu},
  \bibinfo{person}{Jing Yang}, \bibinfo{person}{Lei Feng},
  \bibinfo{person}{Zhijie Ding}, \bibinfo{person}{Yiqiang Chen},
  {et~al\mbox{.}}} \bibinfo{year}{2018}\natexlab{}.
\newblock \showarticletitle{A pervasive approach to EEG-based depression
  detection}.
\newblock \bibinfo{journal}{\emph{Complexity}}  \bibinfo{volume}{2018}
  (\bibinfo{year}{2018}).
\newblock


\bibitem[\protect\citeauthoryear{Canzian and Musolesi}{Canzian and
  Musolesi}{2015}]%
        {canzian2015trajectories}
\bibfield{author}{\bibinfo{person}{Luca Canzian} {and} \bibinfo{person}{Mirco
  Musolesi}.} \bibinfo{year}{2015}\natexlab{}.
\newblock \showarticletitle{Trajectories of depression: unobtrusive monitoring
  of depressive states by means of smartphone mobility traces analysis}. In
  \bibinfo{booktitle}{\emph{Proceedings of the 2015 ACM international joint
  conference on pervasive and ubiquitous computing}}.
  \bibinfo{pages}{1293--1304}.
\newblock


\bibitem[\protect\citeauthoryear{Cao, Lu, and Xu}{Cao et~al\mbox{.}}{2016}]%
        {cao2016deep}
\bibfield{author}{\bibinfo{person}{Shaosheng Cao}, \bibinfo{person}{Wei Lu},
  {and} \bibinfo{person}{Qiongkai Xu}.} \bibinfo{year}{2016}\natexlab{}.
\newblock \showarticletitle{Deep neural networks for learning graph
  representations}. In \bibinfo{booktitle}{\emph{Thirtieth AAAI conference on
  artificial intelligence}}.
\newblock


\bibitem[\protect\citeauthoryear{Che, Xiao, Liang, Jin, Zho, and Wang}{Che
  et~al\mbox{.}}{2017}]%
        {che2017rnn}
\bibfield{author}{\bibinfo{person}{Chao Che}, \bibinfo{person}{Cao Xiao},
  \bibinfo{person}{Jian Liang}, \bibinfo{person}{Bo Jin},
  \bibinfo{person}{Jiayu Zho}, {and} \bibinfo{person}{Fei Wang}.}
  \bibinfo{year}{2017}\natexlab{}.
\newblock \showarticletitle{An rnn architecture with dynamic temporal matching
  for personalized predictions of parkinson's disease}. In
  \bibinfo{booktitle}{\emph{Proceedings of the 2017 SIAM International
  Conference on Data Mining}}. SIAM, \bibinfo{pages}{198--206}.
\newblock


\bibitem[\protect\citeauthoryear{Chen, Yin, Wang, Wang, Nguyen, and Li}{Chen
  et~al\mbox{.}}{2018}]%
        {chen2018pme}
\bibfield{author}{\bibinfo{person}{Hongxu Chen}, \bibinfo{person}{Hongzhi Yin},
  \bibinfo{person}{Weiqing Wang}, \bibinfo{person}{Hao Wang},
  \bibinfo{person}{Quoc Viet~Hung Nguyen}, {and} \bibinfo{person}{Xue Li}.}
  \bibinfo{year}{2018}\natexlab{}.
\newblock \showarticletitle{PME: projected metric embedding on heterogeneous
  networks for link prediction}. In \bibinfo{booktitle}{\emph{Proceedings of
  the 24th ACM SIGKDD International Conference on Knowledge Discovery \&amp;
  Data Mining}}. \bibinfo{pages}{1177--1186}.
\newblock


\bibitem[\protect\citeauthoryear{Cheng, Wang, Zhang, and Hu}{Cheng
  et~al\mbox{.}}{2016}]%
        {cheng2016risk}
\bibfield{author}{\bibinfo{person}{Yu Cheng}, \bibinfo{person}{Fei Wang},
  \bibinfo{person}{Ping Zhang}, {and} \bibinfo{person}{Jianying Hu}.}
  \bibinfo{year}{2016}\natexlab{}.
\newblock \showarticletitle{Risk prediction with electronic health records: A
  deep learning approach}. In \bibinfo{booktitle}{\emph{Proceedings of the 2016
  SIAM International Conference on Data Mining}}. SIAM,
  \bibinfo{pages}{432--440}.
\newblock


\bibitem[\protect\citeauthoryear{Costafreda, Chu, Ashburner, and Fu}{Costafreda
  et~al\mbox{.}}{2009}]%
        {costafreda2009prognostic}
\bibfield{author}{\bibinfo{person}{Sergi~G Costafreda},
  \bibinfo{person}{Carlton Chu}, \bibinfo{person}{John Ashburner}, {and}
  \bibinfo{person}{Cynthia~HY Fu}.} \bibinfo{year}{2009}\natexlab{}.
\newblock \showarticletitle{Prognostic and diagnostic potential of the
  structural neuroanatomy of depression}.
\newblock \bibinfo{journal}{\emph{PloS one}} \bibinfo{volume}{4},
  \bibinfo{number}{7} (\bibinfo{year}{2009}), \bibinfo{pages}{e6353}.
\newblock


\bibitem[\protect\citeauthoryear{Defferrard, Bresson, and
  Vandergheynst}{Defferrard et~al\mbox{.}}{2016}]%
        {defferrard2016convolutional}
\bibfield{author}{\bibinfo{person}{Micha{\"e}l Defferrard},
  \bibinfo{person}{Xavier Bresson}, {and} \bibinfo{person}{Pierre
  Vandergheynst}.} \bibinfo{year}{2016}\natexlab{}.
\newblock \showarticletitle{Convolutional neural networks on graphs with fast
  localized spectral filtering}. In \bibinfo{booktitle}{\emph{Advances in
  neural information processing systems}}. \bibinfo{pages}{3844--3852}.
\newblock


\bibitem[\protect\citeauthoryear{Dong, Chawla, and Swami}{Dong
  et~al\mbox{.}}{2017}]%
        {dong2017metapath2vec}
\bibfield{author}{\bibinfo{person}{Yuxiao Dong}, \bibinfo{person}{Nitesh~V
  Chawla}, {and} \bibinfo{person}{Ananthram Swami}.}
  \bibinfo{year}{2017}\natexlab{}.
\newblock \showarticletitle{metapath2vec: Scalable representation learning for
  heterogeneous networks}. In \bibinfo{booktitle}{\emph{Proceedings of the 23rd
  ACM SIGKDD international conference on knowledge discovery and data mining}}.
  \bibinfo{pages}{135--144}.
\newblock


\bibitem[\protect\citeauthoryear{Fu, Mourao-Miranda, Costafreda, Khanna,
  Marquand, Williams, and Brammer}{Fu et~al\mbox{.}}{2008}]%
        {fu2008pattern}
\bibfield{author}{\bibinfo{person}{Cynthia~HY Fu}, \bibinfo{person}{Janaina
  Mourao-Miranda}, \bibinfo{person}{Sergi~G Costafreda}, \bibinfo{person}{Akash
  Khanna}, \bibinfo{person}{Andre~F Marquand}, \bibinfo{person}{Steve~CR
  Williams}, {and} \bibinfo{person}{Michael~J Brammer}.}
  \bibinfo{year}{2008}\natexlab{}.
\newblock \showarticletitle{Pattern classification of sad facial processing:
  toward the development of neurobiological markers in depression}.
\newblock \bibinfo{journal}{\emph{Biological psychiatry}} \bibinfo{volume}{63},
  \bibinfo{number}{7} (\bibinfo{year}{2008}), \bibinfo{pages}{656--662}.
\newblock


\bibitem[\protect\citeauthoryear{Futoma, Morris, and Lucas}{Futoma
  et~al\mbox{.}}{2015}]%
        {futoma2015comparison}
\bibfield{author}{\bibinfo{person}{Joseph Futoma}, \bibinfo{person}{Jonathan
  Morris}, {and} \bibinfo{person}{Joseph Lucas}.}
  \bibinfo{year}{2015}\natexlab{}.
\newblock \showarticletitle{A comparison of models for predicting early
  hospital readmissions}.
\newblock \bibinfo{journal}{\emph{Journal of biomedical informatics}}
  \bibinfo{volume}{56} (\bibinfo{year}{2015}), \bibinfo{pages}{229--238}.
\newblock


\bibitem[\protect\citeauthoryear{Gallicchio and Micheli}{Gallicchio and
  Micheli}{2010}]%
        {gallicchio2010graph}
\bibfield{author}{\bibinfo{person}{Claudio Gallicchio} {and}
  \bibinfo{person}{Alessio Micheli}.} \bibinfo{year}{2010}\natexlab{}.
\newblock \showarticletitle{Graph echo state networks}. In
  \bibinfo{booktitle}{\emph{The 2010 International Joint Conference on Neural
  Networks (IJCNN)}}. IEEE, \bibinfo{pages}{1--8}.
\newblock


\bibitem[\protect\citeauthoryear{Gao, Zhang, and Xu}{Gao et~al\mbox{.}}{2018}]%
        {gao2018watch}
\bibfield{author}{\bibinfo{person}{Junyu Gao}, \bibinfo{person}{Tianzhu Zhang},
  {and} \bibinfo{person}{Changsheng Xu}.} \bibinfo{year}{2018}\natexlab{}.
\newblock \showarticletitle{Watch, think and attend: End-to-end video
  classification via dynamic knowledge evolution modeling}. In
  \bibinfo{booktitle}{\emph{Proceedings of the 26th ACM international
  conference on Multimedia}}. ACM, \bibinfo{pages}{690--699}.
\newblock


\bibitem[\protect\citeauthoryear{Gao, Zhang, and Xu}{Gao
  et~al\mbox{.}}{2019a}]%
        {gao2019graph}
\bibfield{author}{\bibinfo{person}{Junyu Gao}, \bibinfo{person}{Tianzhu Zhang},
  {and} \bibinfo{person}{Changsheng Xu}.} \bibinfo{year}{2019}\natexlab{a}.
\newblock \showarticletitle{Graph convolutional tracking}. In
  \bibinfo{booktitle}{\emph{Proceedings of the IEEE Conference on Computer
  Vision and Pattern Recognition}}. \bibinfo{pages}{4649--4659}.
\newblock


\bibitem[\protect\citeauthoryear{Gao, Zhang, and Xu}{Gao
  et~al\mbox{.}}{2019b}]%
        {junyu2019AAAI_TS-GCN}
\bibfield{author}{\bibinfo{person}{Junyu Gao}, \bibinfo{person}{Tianzhu Zhang},
  {and} \bibinfo{person}{Changsheng Xu}.} \bibinfo{year}{2019}\natexlab{b}.
\newblock \showarticletitle{I Know the Relationships: Zero-Shot Action
  Recognition via Two-Stream Graph Convolutional Networks and Knowledge
  Graphs}. In \bibinfo{booktitle}{\emph{AAAI}}.
\newblock


\bibitem[\protect\citeauthoryear{Gao, Zhang, and Xu}{Gao et~al\mbox{.}}{2020}]%
        {gao2020learning}
\bibfield{author}{\bibinfo{person}{Junyu Gao}, \bibinfo{person}{Tianzhu Zhang},
  {and} \bibinfo{person}{Changsheng Xu}.} \bibinfo{year}{2020}\natexlab{}.
\newblock \showarticletitle{Learning to Model Relationships for Zero-Shot Video
  Classification}.
\newblock \bibinfo{journal}{\emph{IEEE Transactions on Pattern Analysis and
  Machine Intelligence}} (\bibinfo{year}{2020}).
\newblock


\bibitem[\protect\citeauthoryear{Gilmer, Schoenholz, Riley, Vinyals, and
  Dahl}{Gilmer et~al\mbox{.}}{2017}]%
        {gilmer2017neural}
\bibfield{author}{\bibinfo{person}{Justin Gilmer}, \bibinfo{person}{Samuel~S
  Schoenholz}, \bibinfo{person}{Patrick~F Riley}, \bibinfo{person}{Oriol
  Vinyals}, {and} \bibinfo{person}{George~E Dahl}.}
  \bibinfo{year}{2017}\natexlab{}.
\newblock \showarticletitle{Neural message passing for quantum chemistry}. In
  \bibinfo{booktitle}{\emph{Proceedings of the 34th International Conference on
  Machine Learning-Volume 70}}. JMLR. org, \bibinfo{pages}{1263--1272}.
\newblock


\bibitem[\protect\citeauthoryear{Goel, Saba, Stiber, Whitmire, Fromm, Larson,
  Borriello, and Patel}{Goel et~al\mbox{.}}{2016}]%
        {goel2016spirocall}
\bibfield{author}{\bibinfo{person}{Mayank Goel}, \bibinfo{person}{Elliot Saba},
  \bibinfo{person}{Maia Stiber}, \bibinfo{person}{Eric Whitmire},
  \bibinfo{person}{Josh Fromm}, \bibinfo{person}{Eric~C Larson},
  \bibinfo{person}{Gaetano Borriello}, {and} \bibinfo{person}{Shwetak~N
  Patel}.} \bibinfo{year}{2016}\natexlab{}.
\newblock \showarticletitle{Spirocall: Measuring lung function over a phone
  call}. In \bibinfo{booktitle}{\emph{Proceedings of the 2016 CHI Conference on
  Human Factors in Computing Systems}}. \bibinfo{pages}{5675--5685}.
\newblock


\bibitem[\protect\citeauthoryear{Hahn, Marquand, Ehlis, Dresler,
  Kittel-Schneider, Jarczok, Lesch, Jakob, Mourao-Miranda, Brammer,
  et~al\mbox{.}}{Hahn et~al\mbox{.}}{2011}]%
        {hahn2011integrating}
\bibfield{author}{\bibinfo{person}{Tim Hahn}, \bibinfo{person}{Andre~F
  Marquand}, \bibinfo{person}{Ann-Christine Ehlis}, \bibinfo{person}{Thomas
  Dresler}, \bibinfo{person}{Sarah Kittel-Schneider}, \bibinfo{person}{Tomasz~A
  Jarczok}, \bibinfo{person}{Klaus-Peter Lesch}, \bibinfo{person}{Peter~M
  Jakob}, \bibinfo{person}{Janaina Mourao-Miranda}, \bibinfo{person}{Michael~J
  Brammer}, {et~al\mbox{.}}} \bibinfo{year}{2011}\natexlab{}.
\newblock \showarticletitle{Integrating neurobiological markers of depression}.
\newblock \bibinfo{journal}{\emph{Archives of general psychiatry}}
  \bibinfo{volume}{68}, \bibinfo{number}{4} (\bibinfo{year}{2011}),
  \bibinfo{pages}{361--368}.
\newblock


\bibitem[\protect\citeauthoryear{Henaff, Bruna, and LeCun}{Henaff
  et~al\mbox{.}}{2015}]%
        {henaff2015deep}
\bibfield{author}{\bibinfo{person}{Mikael Henaff}, \bibinfo{person}{Joan
  Bruna}, {and} \bibinfo{person}{Yann LeCun}.} \bibinfo{year}{2015}\natexlab{}.
\newblock \showarticletitle{Deep convolutional networks on graph-structured
  data}.
\newblock \bibinfo{journal}{\emph{arXiv preprint arXiv:1506.05163}}
  (\bibinfo{year}{2015}).
\newblock


\bibitem[\protect\citeauthoryear{Hochreiter and Schmidhuber}{Hochreiter and
  Schmidhuber}{1997}]%
        {hochreiter1997long}
\bibfield{author}{\bibinfo{person}{Sepp Hochreiter} {and}
  \bibinfo{person}{J{\"u}rgen Schmidhuber}.} \bibinfo{year}{1997}\natexlab{}.
\newblock \showarticletitle{Long short-term memory}.
\newblock \bibinfo{journal}{\emph{Neural computation}} \bibinfo{volume}{9},
  \bibinfo{number}{8} (\bibinfo{year}{1997}), \bibinfo{pages}{1735--1780}.
\newblock


\bibitem[\protect\citeauthoryear{Huang, Yang, Gao, Sang, and Xu}{Huang
  et~al\mbox{.}}{2020}]%
        {huang2020knowledge}
\bibfield{author}{\bibinfo{person}{Yi Huang}, \bibinfo{person}{Xiaoshan Yang},
  \bibinfo{person}{Junyu Gao}, \bibinfo{person}{Jitao Sang}, {and}
  \bibinfo{person}{Changsheng Xu}.} \bibinfo{year}{2020}\natexlab{}.
\newblock \showarticletitle{Knowledge-driven Egocentric Multimodal Activity
  Recognition}.
\newblock \bibinfo{journal}{\emph{ACM Transactions on Multimedia Computing,
  Communications, and Applications (TOMM)}} \bibinfo{volume}{16},
  \bibinfo{number}{4} (\bibinfo{year}{2020}), \bibinfo{pages}{1--133}.
\newblock


\bibitem[\protect\citeauthoryear{Jain, Zamir, Savarese, and Saxena}{Jain
  et~al\mbox{.}}{2016}]%
        {jain2016structural}
\bibfield{author}{\bibinfo{person}{Ashesh Jain}, \bibinfo{person}{Amir~R
  Zamir}, \bibinfo{person}{Silvio Savarese}, {and} \bibinfo{person}{Ashutosh
  Saxena}.} \bibinfo{year}{2016}\natexlab{}.
\newblock \showarticletitle{Structural-rnn: Deep learning on spatio-temporal
  graphs}. In \bibinfo{booktitle}{\emph{Proceedings of the ieee conference on
  computer vision and pattern recognition}}. \bibinfo{pages}{5308--5317}.
\newblock


\bibitem[\protect\citeauthoryear{Johnson, Pollard, Shen, Li-Wei, Feng,
  Ghassemi, Moody, Szolovits, Celi, and Mark}{Johnson et~al\mbox{.}}{2016}]%
        {johnson2016mimic}
\bibfield{author}{\bibinfo{person}{Alistair~EW Johnson}, \bibinfo{person}{Tom~J
  Pollard}, \bibinfo{person}{Lu Shen}, \bibinfo{person}{H~Lehman Li-Wei},
  \bibinfo{person}{Mengling Feng}, \bibinfo{person}{Mohammad Ghassemi},
  \bibinfo{person}{Benjamin Moody}, \bibinfo{person}{Peter Szolovits},
  \bibinfo{person}{Leo~Anthony Celi}, {and} \bibinfo{person}{Roger~G Mark}.}
  \bibinfo{year}{2016}\natexlab{}.
\newblock \showarticletitle{MIMIC-III, a freely accessible critical care
  database}.
\newblock \bibinfo{journal}{\emph{Scientific data}} \bibinfo{volume}{3},
  \bibinfo{number}{1} (\bibinfo{year}{2016}), \bibinfo{pages}{1--9}.
\newblock


\bibitem[\protect\citeauthoryear{Kam and Kim}{Kam and Kim}{2017}]%
        {kam2017learning}
\bibfield{author}{\bibinfo{person}{Hye~Jin Kam} {and} \bibinfo{person}{Ha~Young
  Kim}.} \bibinfo{year}{2017}\natexlab{}.
\newblock \showarticletitle{Learning representations for the early detection of
  sepsis with deep neural networks}.
\newblock \bibinfo{journal}{\emph{Computers in biology and medicine}}
  \bibinfo{volume}{89} (\bibinfo{year}{2017}), \bibinfo{pages}{248--255}.
\newblock


\bibitem[\protect\citeauthoryear{Kho, Hayes, Rasmussen-Torvik, Pacheco,
  Thompson, Armstrong, Denny, Peissig, Miller, Wei, et~al\mbox{.}}{Kho
  et~al\mbox{.}}{2012}]%
        {kho2012use}
\bibfield{author}{\bibinfo{person}{Abel~N Kho}, \bibinfo{person}{M~Geoffrey
  Hayes}, \bibinfo{person}{Laura Rasmussen-Torvik}, \bibinfo{person}{Jennifer~A
  Pacheco}, \bibinfo{person}{William~K Thompson}, \bibinfo{person}{Loren~L
  Armstrong}, \bibinfo{person}{Joshua~C Denny}, \bibinfo{person}{Peggy~L
  Peissig}, \bibinfo{person}{Aaron~W Miller}, \bibinfo{person}{Wei-Qi Wei},
  {et~al\mbox{.}}} \bibinfo{year}{2012}\natexlab{}.
\newblock \showarticletitle{Use of diverse electronic medical record systems to
  identify genetic risk for type 2 diabetes within a genome-wide association
  study}.
\newblock \bibinfo{journal}{\emph{Journal of the American Medical Informatics
  Association}} \bibinfo{volume}{19}, \bibinfo{number}{2}
  (\bibinfo{year}{2012}), \bibinfo{pages}{212--218}.
\newblock


\bibitem[\protect\citeauthoryear{Kipf and Welling}{Kipf and Welling}{2016}]%
        {kipf2016semi}
\bibfield{author}{\bibinfo{person}{Thomas~N Kipf} {and} \bibinfo{person}{Max
  Welling}.} \bibinfo{year}{2016}\natexlab{}.
\newblock \showarticletitle{Semi-supervised classification with graph
  convolutional networks}.
\newblock \bibinfo{journal}{\emph{arXiv preprint arXiv:1609.02907}}
  (\bibinfo{year}{2016}).
\newblock


\bibitem[\protect\citeauthoryear{Koenig, Seeck, Eckstein, Mainka, Huebner,
  Voss, and Weber}{Koenig et~al\mbox{.}}{2016}]%
        {koenig2016validation}
\bibfield{author}{\bibinfo{person}{Nicole Koenig}, \bibinfo{person}{Andrea
  Seeck}, \bibinfo{person}{Jens Eckstein}, \bibinfo{person}{Andreas Mainka},
  \bibinfo{person}{Thomas Huebner}, \bibinfo{person}{Andreas Voss}, {and}
  \bibinfo{person}{Stefan Weber}.} \bibinfo{year}{2016}\natexlab{}.
\newblock \showarticletitle{Validation of a new heart rate measurement
  algorithm for fingertip recording of video signals with smartphones}.
\newblock \bibinfo{journal}{\emph{Telemedicine and e-Health}}
  \bibinfo{volume}{22}, \bibinfo{number}{8} (\bibinfo{year}{2016}),
  \bibinfo{pages}{631--636}.
\newblock


\bibitem[\protect\citeauthoryear{Laaksonen and Oja}{Laaksonen and Oja}{2002}]%
        {Laaksonen2002Classification}
\bibfield{author}{\bibinfo{person}{Jorma Laaksonen} {and}
  \bibinfo{person}{Erkki Oja}.} \bibinfo{year}{2002}\natexlab{}.
\newblock \showarticletitle{Classification with Learning k-Nearest Neighbors}.
  In \bibinfo{booktitle}{\emph{Proceedings of International Conference on
  Neural Networks (ICNN'96)}}.
\newblock


\bibitem[\protect\citeauthoryear{Lane, Mohammod, Lin, Yang, Lu, Ali, Doryab,
  Berke, Choudhury, and Campbell}{Lane et~al\mbox{.}}{2011}]%
        {lane2011bewell}
\bibfield{author}{\bibinfo{person}{Nicholas~D Lane}, \bibinfo{person}{Mashfiqui
  Mohammod}, \bibinfo{person}{Mu Lin}, \bibinfo{person}{Xiaochao Yang},
  \bibinfo{person}{Hong Lu}, \bibinfo{person}{Shahid Ali},
  \bibinfo{person}{Afsaneh Doryab}, \bibinfo{person}{Ethan Berke},
  \bibinfo{person}{Tanzeem Choudhury}, {and} \bibinfo{person}{Andrew
  Campbell}.} \bibinfo{year}{2011}\natexlab{}.
\newblock \showarticletitle{Bewell: A smartphone application to monitor, model
  and promote wellbeing}. In \bibinfo{booktitle}{\emph{5th international ICST
  conference on pervasive computing technologies for healthcare}}.
  \bibinfo{pages}{23--26}.
\newblock


\bibitem[\protect\citeauthoryear{LeCun, Bengio, and Hinton}{LeCun
  et~al\mbox{.}}{2015}]%
        {lecun2015deep}
\bibfield{author}{\bibinfo{person}{Yann LeCun}, \bibinfo{person}{Yoshua
  Bengio}, {and} \bibinfo{person}{Geoffrey Hinton}.}
  \bibinfo{year}{2015}\natexlab{}.
\newblock \showarticletitle{Deep learning}.
\newblock \bibinfo{journal}{\emph{nature}} \bibinfo{volume}{521},
  \bibinfo{number}{7553} (\bibinfo{year}{2015}), \bibinfo{pages}{436--444}.
\newblock


\bibitem[\protect\citeauthoryear{Li and Trocan}{Li and Trocan}{2019}]%
        {li2019deep}
\bibfield{author}{\bibinfo{person}{Honggui Li} {and} \bibinfo{person}{Maria
  Trocan}.} \bibinfo{year}{2019}\natexlab{}.
\newblock \showarticletitle{Deep learning of smartphone sensor data for
  personal health assistance}.
\newblock \bibinfo{journal}{\emph{Microelectronics Journal}}
  \bibinfo{volume}{88} (\bibinfo{year}{2019}), \bibinfo{pages}{164--172}.
\newblock


\bibitem[\protect\citeauthoryear{Li, Yu, Shahabi, and Liu}{Li
  et~al\mbox{.}}{2017}]%
        {li2017diffusion}
\bibfield{author}{\bibinfo{person}{Yaguang Li}, \bibinfo{person}{Rose Yu},
  \bibinfo{person}{Cyrus Shahabi}, {and} \bibinfo{person}{Yan Liu}.}
  \bibinfo{year}{2017}\natexlab{}.
\newblock \showarticletitle{Diffusion convolutional recurrent neural network:
  Data-driven traffic forecasting}.
\newblock \bibinfo{journal}{\emph{arXiv preprint arXiv:1707.01926}}
  (\bibinfo{year}{2017}).
\newblock


\bibitem[\protect\citeauthoryear{Machado, Gomes, Gamboa, Paix{\~a}o, and
  Costa}{Machado et~al\mbox{.}}{2015}]%
        {machado2015human}
\bibfield{author}{\bibinfo{person}{In{\^e}s~P Machado},
  \bibinfo{person}{A~Luisa Gomes}, \bibinfo{person}{Hugo Gamboa},
  \bibinfo{person}{V{\'\i}tor Paix{\~a}o}, {and} \bibinfo{person}{Rui~M
  Costa}.} \bibinfo{year}{2015}\natexlab{}.
\newblock \showarticletitle{Human activity data discovery from triaxial
  accelerometer sensor: Non-supervised learning sensitivity to feature
  extraction parametrization}.
\newblock \bibinfo{journal}{\emph{Information Processing \&amp; Management}}
  \bibinfo{volume}{51}, \bibinfo{number}{2} (\bibinfo{year}{2015}),
  \bibinfo{pages}{204--214}.
\newblock


\bibitem[\protect\citeauthoryear{machado2015human, Oren, Chen, Dai~AM, Liu,
  et~al\mbox{.}}{machado2015human et~al\mbox{.}}{1801}]%
        {rajkomar1801scalable}
\bibfield{author}{\bibinfo{person}{A machado2015human}, \bibinfo{person}{E
  Oren}, \bibinfo{person}{K Chen}, \bibinfo{person}{Hajaj~N Dai~AM},
  \bibinfo{person}{PJ Liu}, {et~al\mbox{.}}} \bibinfo{year}{1801}\natexlab{}.
\newblock \bibinfo{title}{Scalable and accurate deep learning for electronic
  health records. npj Digit Med. 2018}.
\newblock
\newblock


\bibitem[\protect\citeauthoryear{McGrath, Kelley, Holtzheimer, Dunlop,
  Craighead, Franco, Craddock, and Mayberg}{McGrath et~al\mbox{.}}{2013}]%
        {mcgrath2013toward}
\bibfield{author}{\bibinfo{person}{Callie~L McGrath}, \bibinfo{person}{Mary~E
  Kelley}, \bibinfo{person}{Paul~E Holtzheimer}, \bibinfo{person}{Boadie~W
  Dunlop}, \bibinfo{person}{W~Edward Craighead}, \bibinfo{person}{Alexandre~R
  Franco}, \bibinfo{person}{R~Cameron Craddock}, {and} \bibinfo{person}{Helen~S
  Mayberg}.} \bibinfo{year}{2013}\natexlab{}.
\newblock \showarticletitle{Toward a neuroimaging treatment selection biomarker
  for major depressive disorder}.
\newblock \bibinfo{journal}{\emph{JAMA psychiatry}} \bibinfo{volume}{70},
  \bibinfo{number}{8} (\bibinfo{year}{2013}), \bibinfo{pages}{821--829}.
\newblock


\bibitem[\protect\citeauthoryear{Min, Doryab, Wiese, Amini, Zimmerman, and
  Hong}{Min et~al\mbox{.}}{2014}]%
        {min2014toss}
\bibfield{author}{\bibinfo{person}{Jun-Ki Min}, \bibinfo{person}{Afsaneh
  Doryab}, \bibinfo{person}{Jason Wiese}, \bibinfo{person}{Shahriyar Amini},
  \bibinfo{person}{John Zimmerman}, {and} \bibinfo{person}{Jason~I Hong}.}
  \bibinfo{year}{2014}\natexlab{}.
\newblock \showarticletitle{Toss'n'turn: smartphone as sleep and sleep quality
  detector}. In \bibinfo{booktitle}{\emph{Proceedings of the SIGCHI conference
  on human factors in computing systems}}. \bibinfo{pages}{477--486}.
\newblock


\bibitem[\protect\citeauthoryear{Min, Bao, Mei, Zhu, Rui, and Jiang}{Min
  et~al\mbox{.}}{2017}]%
        {min2017you}
\bibfield{author}{\bibinfo{person}{Weiqing Min}, \bibinfo{person}{Bing-Kun
  Bao}, \bibinfo{person}{Shuhuan Mei}, \bibinfo{person}{Yaohui Zhu},
  \bibinfo{person}{Yong Rui}, {and} \bibinfo{person}{Shuqiang Jiang}.}
  \bibinfo{year}{2017}\natexlab{}.
\newblock \showarticletitle{You are what you eat: Exploring rich recipe
  information for cross-region food analysis}.
\newblock \bibinfo{journal}{\emph{IEEE Transactions on Multimedia}}
  \bibinfo{volume}{20}, \bibinfo{number}{4} (\bibinfo{year}{2017}),
  \bibinfo{pages}{950--964}.
\newblock


\bibitem[\protect\citeauthoryear{Mwangi, Matthews, and Steele}{Mwangi
  et~al\mbox{.}}{2012}]%
        {mwangi2012prediction}
\bibfield{author}{\bibinfo{person}{Benson Mwangi}, \bibinfo{person}{Keith
  Matthews}, {and} \bibinfo{person}{J~Douglas Steele}.}
  \bibinfo{year}{2012}\natexlab{}.
\newblock \showarticletitle{Prediction of illness severity in patients with
  major depression using structural MR brain scans}.
\newblock \bibinfo{journal}{\emph{Journal of Magnetic Resonance Imaging}}
  \bibinfo{volume}{35}, \bibinfo{number}{1} (\bibinfo{year}{2012}),
  \bibinfo{pages}{64--71}.
\newblock


\bibitem[\protect\citeauthoryear{Nag, Pandey, Putzel, Bhimaraju, Krishnan, and
  Jain}{Nag et~al\mbox{.}}{2018}]%
        {nag2018cross}
\bibfield{author}{\bibinfo{person}{Nitish Nag}, \bibinfo{person}{Vaibhav
  Pandey}, \bibinfo{person}{Preston~J Putzel}, \bibinfo{person}{Hari
  Bhimaraju}, \bibinfo{person}{Srikanth Krishnan}, {and}
  \bibinfo{person}{Ramesh Jain}.} \bibinfo{year}{2018}\natexlab{}.
\newblock \showarticletitle{Cross-modal health state estimation}. In
  \bibinfo{booktitle}{\emph{Proceedings of the 26th ACM international
  conference on Multimedia}}. \bibinfo{pages}{1993--2002}.
\newblock


\bibitem[\protect\citeauthoryear{Nandakumar, Gollakota, and Watson}{Nandakumar
  et~al\mbox{.}}{2015}]%
        {nandakumar2015contactless}
\bibfield{author}{\bibinfo{person}{Rajalakshmi Nandakumar},
  \bibinfo{person}{Shyamnath Gollakota}, {and} \bibinfo{person}{Nathaniel
  Watson}.} \bibinfo{year}{2015}\natexlab{}.
\newblock \showarticletitle{Contactless sleep apnea detection on smartphones}.
  In \bibinfo{booktitle}{\emph{Proceedings of the 13th annual international
  conference on mobile systems, applications, and services}}.
  \bibinfo{pages}{45--57}.
\newblock


\bibitem[\protect\citeauthoryear{Natale, Drejak, Erbacci, Tonetti, Fabbri, and
  Martoni}{Natale et~al\mbox{.}}{2012}]%
        {natale2012monitoring}
\bibfield{author}{\bibinfo{person}{Vincenzo Natale}, \bibinfo{person}{Maciek
  Drejak}, \bibinfo{person}{Alex Erbacci}, \bibinfo{person}{Lorenzo Tonetti},
  \bibinfo{person}{Marco Fabbri}, {and} \bibinfo{person}{Monica Martoni}.}
  \bibinfo{year}{2012}\natexlab{}.
\newblock \showarticletitle{Monitoring sleep with a smartphone accelerometer}.
\newblock \bibinfo{journal}{\emph{Sleep and Biological Rhythms}}
  \bibinfo{volume}{10}, \bibinfo{number}{4} (\bibinfo{year}{2012}),
  \bibinfo{pages}{287--292}.
\newblock


\bibitem[\protect\citeauthoryear{Niepert, Ahmed, and Kutzkov}{Niepert
  et~al\mbox{.}}{2016}]%
        {niepert2016learning}
\bibfield{author}{\bibinfo{person}{Mathias Niepert}, \bibinfo{person}{Mohamed
  Ahmed}, {and} \bibinfo{person}{Konstantin Kutzkov}.}
  \bibinfo{year}{2016}\natexlab{}.
\newblock \showarticletitle{Learning convolutional neural networks for graphs}.
  In \bibinfo{booktitle}{\emph{International conference on machine learning}}.
  \bibinfo{pages}{2014--2023}.
\newblock


\bibitem[\protect\citeauthoryear{Peissig, Rasmussen, Berg, Linneman, McCarty,
  Waudby, Chen, Denny, Wilke, Pathak, et~al\mbox{.}}{Peissig
  et~al\mbox{.}}{2012}]%
        {peissig2012importance}
\bibfield{author}{\bibinfo{person}{Peggy~L Peissig}, \bibinfo{person}{Luke~V
  Rasmussen}, \bibinfo{person}{Richard~L Berg}, \bibinfo{person}{James~G
  Linneman}, \bibinfo{person}{Catherine~A McCarty}, \bibinfo{person}{Carol
  Waudby}, \bibinfo{person}{Lin Chen}, \bibinfo{person}{Joshua~C Denny},
  \bibinfo{person}{Russell~A Wilke}, \bibinfo{person}{Jyotishman Pathak},
  {et~al\mbox{.}}} \bibinfo{year}{2012}\natexlab{}.
\newblock \showarticletitle{Importance of multi-modal approaches to effectively
  identify cataract cases from electronic health records}.
\newblock \bibinfo{journal}{\emph{Journal of the American Medical Informatics
  Association}} \bibinfo{volume}{19}, \bibinfo{number}{2}
  (\bibinfo{year}{2012}), \bibinfo{pages}{225--234}.
\newblock


\bibitem[\protect\citeauthoryear{Pennington, Socher, and Manning}{Pennington
  et~al\mbox{.}}{2014}]%
        {pennington2014glove}
\bibfield{author}{\bibinfo{person}{Jeffrey Pennington},
  \bibinfo{person}{Richard Socher}, {and} \bibinfo{person}{Christopher~D
  Manning}.} \bibinfo{year}{2014}\natexlab{}.
\newblock \showarticletitle{Glove: Global vectors for word representation}. In
  \bibinfo{booktitle}{\emph{Proceedings of the 2014 conference on empirical
  methods in natural language processing (EMNLP)}}.
  \bibinfo{pages}{1532--1543}.
\newblock


\bibitem[\protect\citeauthoryear{Pollak, Adams, and Gay}{Pollak
  et~al\mbox{.}}{2011}]%
        {pollak2011pam}
\bibfield{author}{\bibinfo{person}{John~P Pollak}, \bibinfo{person}{Phil
  Adams}, {and} \bibinfo{person}{Geri Gay}.} \bibinfo{year}{2011}\natexlab{}.
\newblock \showarticletitle{PAM: a photographic affect meter for frequent, in
  situ measurement of affect}. In \bibinfo{booktitle}{\emph{Proceedings of the
  SIGCHI conference on Human factors in computing systems}}.
  \bibinfo{pages}{725--734}.
\newblock


\bibitem[\protect\citeauthoryear{Qi, Yang, and Xu}{Qi et~al\mbox{.}}{2020}]%
        {qi2020emotion}
\bibfield{author}{\bibinfo{person}{Fan Qi}, \bibinfo{person}{Xiaoshan Yang},
  {and} \bibinfo{person}{Changsheng Xu}.} \bibinfo{year}{2020}\natexlab{}.
\newblock \showarticletitle{Emotion Knowledge Driven Video Highlight
  Detection}.
\newblock \bibinfo{journal}{\emph{IEEE Transactions on Multimedia}}
  (\bibinfo{year}{2020}).
\newblock


\bibitem[\protect\citeauthoryear{Ridge, Wadsworth, Miller, Saykin, Green,
  Kauwe, Initiative, et~al\mbox{.}}{Ridge et~al\mbox{.}}{2018}]%
        {ridge2018assembly}
\bibfield{author}{\bibinfo{person}{Perry~G Ridge}, \bibinfo{person}{Mark~E
  Wadsworth}, \bibinfo{person}{Justin~B Miller}, \bibinfo{person}{Andrew~J
  Saykin}, \bibinfo{person}{Robert~C Green}, \bibinfo{person}{John~SK Kauwe},
  \bibinfo{person}{Alzheimer's Disease~Neuroimaging Initiative},
  {et~al\mbox{.}}} \bibinfo{year}{2018}\natexlab{}.
\newblock \showarticletitle{Assembly of 809 whole mitochondrial genomes with
  clinical, imaging, and fluid biomarker phenotyping}.
\newblock \bibinfo{journal}{\emph{Alzheimer's \& Dementia}}
  \bibinfo{volume}{14}, \bibinfo{number}{4} (\bibinfo{year}{2018}),
  \bibinfo{pages}{514--519}.
\newblock


\bibitem[\protect\citeauthoryear{Rosa, Portugal, Hahn, Fallgatter, Garrido,
  Shawe-Taylor, and Mourao-Miranda}{Rosa et~al\mbox{.}}{2015}]%
        {rosa2015sparse}
\bibfield{author}{\bibinfo{person}{Maria~J Rosa}, \bibinfo{person}{Liana
  Portugal}, \bibinfo{person}{Tim Hahn}, \bibinfo{person}{Andreas~J
  Fallgatter}, \bibinfo{person}{Marta~I Garrido}, \bibinfo{person}{John
  Shawe-Taylor}, {and} \bibinfo{person}{Janaina Mourao-Miranda}.}
  \bibinfo{year}{2015}\natexlab{}.
\newblock \showarticletitle{Sparse network-based models for patient
  classification using fMRI}.
\newblock \bibinfo{journal}{\emph{Neuroimage}}  \bibinfo{volume}{105}
  (\bibinfo{year}{2015}), \bibinfo{pages}{493--506}.
\newblock


\bibitem[\protect\citeauthoryear{Scarselli, Gori, Tsoi, Hagenbuchner, and
  Monfardini}{Scarselli et~al\mbox{.}}{2008}]%
        {scarselli2008graph}
\bibfield{author}{\bibinfo{person}{Franco Scarselli}, \bibinfo{person}{Marco
  Gori}, \bibinfo{person}{Ah~Chung Tsoi}, \bibinfo{person}{Markus
  Hagenbuchner}, {and} \bibinfo{person}{Gabriele Monfardini}.}
  \bibinfo{year}{2008}\natexlab{}.
\newblock \showarticletitle{The graph neural network model}.
\newblock \bibinfo{journal}{\emph{IEEE Transactions on Neural Networks}}
  \bibinfo{volume}{20}, \bibinfo{number}{1} (\bibinfo{year}{2008}),
  \bibinfo{pages}{61--80}.
\newblock


\bibitem[\protect\citeauthoryear{Seo, Defferrard, Vandergheynst, and
  Bresson}{Seo et~al\mbox{.}}{2018}]%
        {seo2018structured}
\bibfield{author}{\bibinfo{person}{Youngjoo Seo}, \bibinfo{person}{Micha{\"e}l
  Defferrard}, \bibinfo{person}{Pierre Vandergheynst}, {and}
  \bibinfo{person}{Xavier Bresson}.} \bibinfo{year}{2018}\natexlab{}.
\newblock \showarticletitle{Structured sequence modeling with graph
  convolutional recurrent networks}. In \bibinfo{booktitle}{\emph{International
  Conference on Neural Information Processing}}. Springer,
  \bibinfo{pages}{362--373}.
\newblock


\bibitem[\protect\citeauthoryear{Stafford, Lin, and Xu}{Stafford
  et~al\mbox{.}}{2016}]%
        {stafford2016flappy}
\bibfield{author}{\bibinfo{person}{Matthew Stafford}, \bibinfo{person}{Feng
  Lin}, {and} \bibinfo{person}{Wenyao Xu}.} \bibinfo{year}{2016}\natexlab{}.
\newblock \showarticletitle{Flappy Breath: A Smartphone-Based Breath Exergame}.
  In \bibinfo{booktitle}{\emph{2016 IEEE First International Conference on
  Connected Health: Applications, Systems and Engineering Technologies
  (CHASE)}}. IEEE, \bibinfo{pages}{332--333}.
\newblock


\bibitem[\protect\citeauthoryear{Sun and Han}{Sun and Han}{2013}]%
        {sun2013mining}
\bibfield{author}{\bibinfo{person}{Yizhou Sun} {and} \bibinfo{person}{Jiawei
  Han}.} \bibinfo{year}{2013}\natexlab{}.
\newblock \showarticletitle{Mining heterogeneous information networks: a
  structural analysis approach}.
\newblock \bibinfo{journal}{\emph{Acm Sigkdd Explorations Newsletter}}
  \bibinfo{volume}{14}, \bibinfo{number}{2} (\bibinfo{year}{2013}),
  \bibinfo{pages}{20--28}.
\newblock


\bibitem[\protect\citeauthoryear{{Tin Kam Ho}}{{Tin Kam Ho}}{1995}]%
        {598994}
\bibfield{author}{\bibinfo{person}{{Tin Kam Ho}}.}
  \bibinfo{year}{1995}\natexlab{}.
\newblock \showarticletitle{Random decision forests}. In
  \bibinfo{booktitle}{\emph{Proceedings of 3rd International Conference on
  Document Analysis and Recognition}}, Vol.~\bibinfo{volume}{1}.
  \bibinfo{pages}{278--282 vol.1}.
\newblock


\bibitem[\protect\citeauthoryear{Vaswani, Shazeer, Parmar, Uszkoreit, Jones,
  Gomez, Kaiser, and Polosukhin}{Vaswani et~al\mbox{.}}{2017}]%
        {vaswani2017attention}
\bibfield{author}{\bibinfo{person}{Ashish Vaswani}, \bibinfo{person}{Noam
  Shazeer}, \bibinfo{person}{Niki Parmar}, \bibinfo{person}{Jakob Uszkoreit},
  \bibinfo{person}{Llion Jones}, \bibinfo{person}{Aidan~N Gomez},
  \bibinfo{person}{{\L}ukasz Kaiser}, {and} \bibinfo{person}{Illia
  Polosukhin}.} \bibinfo{year}{2017}\natexlab{}.
\newblock \showarticletitle{Attention is all you need}. In
  \bibinfo{booktitle}{\emph{Advances in neural information processing
  systems}}. \bibinfo{pages}{5998--6008}.
\newblock


\bibitem[\protect\citeauthoryear{Wang, Cui, and Zhu}{Wang
  et~al\mbox{.}}{2016}]%
        {wang2016structural}
\bibfield{author}{\bibinfo{person}{Daixin Wang}, \bibinfo{person}{Peng Cui},
  {and} \bibinfo{person}{Wenwu Zhu}.} \bibinfo{year}{2016}\natexlab{}.
\newblock \showarticletitle{Structural deep network embedding}. In
  \bibinfo{booktitle}{\emph{Proceedings of the 22nd ACM SIGKDD international
  conference on Knowledge discovery and data mining}}.
  \bibinfo{pages}{1225--1234}.
\newblock


\bibitem[\protect\citeauthoryear{Wang, Chen, Chen, Li, Harari, Tignor, Zhou,
  Ben-Zeev, and Campbell}{Wang et~al\mbox{.}}{2014}]%
        {wang2014studentlife}
\bibfield{author}{\bibinfo{person}{Rui Wang}, \bibinfo{person}{Fanglin Chen},
  \bibinfo{person}{Zhenyu Chen}, \bibinfo{person}{Tianxing Li},
  \bibinfo{person}{Gabriella Harari}, \bibinfo{person}{Stefanie Tignor},
  \bibinfo{person}{Xia Zhou}, \bibinfo{person}{Dror Ben-Zeev}, {and}
  \bibinfo{person}{Andrew~T Campbell}.} \bibinfo{year}{2014}\natexlab{}.
\newblock \showarticletitle{StudentLife: assessing mental health, academic
  performance and behavioral trends of college students using smartphones}. In
  \bibinfo{booktitle}{\emph{Proceedings of the 2014 ACM international joint
  conference on pervasive and ubiquitous computing}}. \bibinfo{pages}{3--14}.
\newblock


\bibitem[\protect\citeauthoryear{Wang, Ji, Shi, Wang, Ye, Cui, and Yu}{Wang
  et~al\mbox{.}}{2019}]%
        {wang2019heterogeneous}
\bibfield{author}{\bibinfo{person}{Xiao Wang}, \bibinfo{person}{Houye Ji},
  \bibinfo{person}{Chuan Shi}, \bibinfo{person}{Bai Wang},
  \bibinfo{person}{Yanfang Ye}, \bibinfo{person}{Peng Cui}, {and}
  \bibinfo{person}{Philip~S Yu}.} \bibinfo{year}{2019}\natexlab{}.
\newblock \showarticletitle{Heterogeneous graph attention network}. In
  \bibinfo{booktitle}{\emph{The World Wide Web Conference}}.
  \bibinfo{pages}{2022--2032}.
\newblock


\bibitem[\protect\citeauthoryear{Watson, Clark, and Tellegen}{Watson
  et~al\mbox{.}}{1988}]%
        {watson1988development}
\bibfield{author}{\bibinfo{person}{David Watson}, \bibinfo{person}{Lee~Anna
  Clark}, {and} \bibinfo{person}{Auke Tellegen}.}
  \bibinfo{year}{1988}\natexlab{}.
\newblock \showarticletitle{Development and validation of brief measures of
  positive and negative affect: the PANAS scales.}
\newblock \bibinfo{journal}{\emph{Journal of personality and social
  psychology}} \bibinfo{volume}{54}, \bibinfo{number}{6}
  (\bibinfo{year}{1988}), \bibinfo{pages}{1063}.
\newblock


\bibitem[\protect\citeauthoryear{Wu, Pan, Chen, Long, Zhang, and Philip}{Wu
  et~al\mbox{.}}{2020}]%
        {wu2020comprehensive}
\bibfield{author}{\bibinfo{person}{Zonghan Wu}, \bibinfo{person}{Shirui Pan},
  \bibinfo{person}{Fengwen Chen}, \bibinfo{person}{Guodong Long},
  \bibinfo{person}{Chengqi Zhang}, {and} \bibinfo{person}{S~Yu Philip}.}
  \bibinfo{year}{2020}\natexlab{}.
\newblock \showarticletitle{A comprehensive survey on graph neural networks}.
\newblock \bibinfo{journal}{\emph{IEEE Transactions on Neural Networks and
  Learning Systems}} (\bibinfo{year}{2020}).
\newblock


\bibitem[\protect\citeauthoryear{Yebda, Benois-Pineau, Amieva, and
  Frolicher}{Yebda et~al\mbox{.}}{2019}]%
        {yebda2019multi}
\bibfield{author}{\bibinfo{person}{Thinhinane Yebda}, \bibinfo{person}{Jenny
  Benois-Pineau}, \bibinfo{person}{Helene Amieva}, {and}
  \bibinfo{person}{Benjamin Frolicher}.} \bibinfo{year}{2019}\natexlab{}.
\newblock \showarticletitle{Multi-sensing of fragile persons for risk situation
  detection: devices, methods, challenges}. In \bibinfo{booktitle}{\emph{2019
  International Conference on Content-Based Multimedia Indexing (CBMI)}}. IEEE,
  \bibinfo{pages}{1--6}.
\newblock


\bibitem[\protect\citeauthoryear{Zhang, Huang, Yu, Yang, Wang, and Sang}{Zhang
  et~al\mbox{.}}{2019}]%
        {zhang2019multimodal}
\bibfield{author}{\bibinfo{person}{Weiming Zhang}, \bibinfo{person}{Yi Huang},
  \bibinfo{person}{Wanting Yu}, \bibinfo{person}{Xiaoshan Yang},
  \bibinfo{person}{Wei Wang}, {and} \bibinfo{person}{Jitao Sang}.}
  \bibinfo{year}{2019}\natexlab{}.
\newblock \showarticletitle{Multimodal attribute and feature embedding for
  activity recognition}.
\newblock In \bibinfo{booktitle}{\emph{Proceedings of the ACM Multimedia
  Asia}}. \bibinfo{pages}{1--7}.
\newblock


\bibitem[\protect\citeauthoryear{Zhang, Cui, and Zhu}{Zhang
  et~al\mbox{.}}{2020}]%
        {zhang2020deep}
\bibfield{author}{\bibinfo{person}{Ziwei Zhang}, \bibinfo{person}{Peng Cui},
  {and} \bibinfo{person}{Wenwu Zhu}.} \bibinfo{year}{2020}\natexlab{}.
\newblock \showarticletitle{Deep learning on graphs: A survey}.
\newblock \bibinfo{journal}{\emph{IEEE Transactions on Knowledge and Data
  Engineering}} (\bibinfo{year}{2020}).
\newblock


\end{thebibliography}

\end{document}